\documentclass[10pt,journal,compsoc]{IEEEtran}
\usepackage[switch]{lineno}

\usepackage{booktabs}

\ifCLASSOPTIONcompsoc
  \usepackage[nocompress]{cite}
\else
  \usepackage{cite}
\fi
\ifCLASSINFOpdf
   \usepackage[pdftex]{graphicx}
   \usepackage{wrapfig}
   \usepackage{xcolor}

\else
\fi
\usepackage{amsmath}
\usepackage{algorithmic}
\usepackage{array}
\ifCLASSOPTIONcompsoc
 \usepackage[caption=false,font=footnotesize,labelfont=sf,textfont=sf]{subfig}
\else
 \usepackage[caption=false,font=footnotesize]{subfig}
\fi
\usepackage{url}
\usepackage{hyperref}

\begin{document}
% \linenumbers

\title{Robust Reflection Removal with Flash-only Cues in the Wild}
% \title{Deep Video Prior}


\author{Chenyang~Lei$^*$,
	Xudong~Jiang$^*$\thanks{$^*$ Joint first authors},
	and~Qifeng~Chen,~\IEEEmembership{Member,~IEEE}% <-this % stops a space

	% \author{Chenyang~Lei, ~\IEEEmembership{Member,~IEEE,}
	%         Yazhou~Xing, ~\IEEEmembership{Member,~IEEE,}
	%         Hao~Ouyang,~\IEEEmembership{Member,~IEEE,}
	%         and~Qifeng~Chen,~\IEEEmembership{Member,~IEEE}% <-this % stops a space
	\IEEEcompsocitemizethanks{\IEEEcompsocthanksitem C. Lei, X. Jiang, and Q. Chen (corresponding author) are with the Department of Computer Science and Engineering, The Hong Kong University of Science and Technology, Clear Water Bay, Hong
		Kong, China.\protect\\
		Email: \{cleiaa, xjiangan\}@connect.ust.hk, cqf@ust.hk

		%E-mail: see http://www.michaelshell.org/contact.html
		% \IEEEcompsocthanksitem J. Doe and J. Doe are with Anonymous University.
	}% <-this % stops an unwanted space
}


\markboth{%Journal of \LaTeX\ Class Files,~Vol.~14, No.~8, August~2015
}
{Shell \MakeLowercase{\textit{et al.}}: Bare Demo of IEEEtran.cls for Computer Society Journals}
\IEEEtitleabstractindextext{%
	\begin{abstract}
	We propose a simple yet effective reflection-free cue for robust reflection removal from a pair of flash and ambient (no-flash) images. The reflection-free cue exploits a flash-only image obtained by subtracting the ambient image from the corresponding flash image in raw data space. The flash-only image is equivalent to an image taken in a dark environment with only a flash on. This flash-only image is visually reflection-free and thus can provide robust cues to infer the reflection in the ambient image. Since the flash-only image usually has artifacts, we further propose a dedicated model that not only utilizes the reflection-free cue but also avoids introducing artifacts, which helps accurately estimate reflection and transmission. Our experiments on real-world images with various types of reflection demonstrate the effectiveness of our model with reflection-free flash-only cues: our model outperforms state-of-the-art reflection removal approaches by more than 5.23dB in PSNR. We extend our approach to handheld photography to address the misalignment between the flash and no-flash pair. With misaligned training data and the alignment module, our aligned model outperforms our previous version by more than 3.19dB in PSNR on a misaligned dataset. We also study using linear RGB images as training data. Our source code and dataset are publicly available at \url{https://github.com/ChenyangLEI/flash-reflection-removal}.
		%We propose a simple yet effective reflection-free cue for robust reflection removal from a pair of flash and ambient (no-flash) images. The reflection-free cue exploits a flash-only image obtained by subtracting the ambient image from the corresponding flash image in raw data space. The flash-only image is equivalent to an image taken in a dark environment with only a flash on. We observe that this flash-only image is visually reflection-free, and thus it can provide robust cues to infer the reflection in the ambient image. Since the flash-only image usually has artifacts, we further propose a dedicated model that not only utilizes the reflection-free cue but also avoids introducing artifacts, which helps accurately estimate reflection and transmission. Our experiments on real-world images with various types of reflection demonstrate the effectiveness of our model with reflection-free flash-only cues: our model outperforms state-of-the-art reflection removal approaches by more than 5.23dB in PSNR, 0.04 in SSIM, and 0.068 in LPIPS. We further extend our approach to handheld photography that can handle the misalignment between the flash and no-flash pair. With misaligned training data and the alignment module, our aligned model outperforms our previous version by more than 3.19dB in PSNR, 0.039 in SSIM, and 0.028 in LPIPS on a misaligned dataset. We also study the performance of using linear RGB images as training data. Our source code and dataset are publicly available at \url{https://github.com/ChenyangLEI/flash-reflection-removal}.
	\end{abstract}

	}

\maketitle

\IEEEdisplaynontitleabstractindextext

% \fi
\IEEEpeerreviewmaketitle

\IEEEraisesectionheading{\section{Introduction}\label{sec:introduction}}

\IEEEPARstart{A}{n} image taken by a camera in front of a glass often contains undesirable reflections. In the process of image formation with reflection, the irradiance received by a camera can be approximately modeled as the sum of transmission and reflection. In this paper, we are interested in recovering a clear transmission image by removing reflection from the ambient image (captured image). Reflection removal is an important application in computational photography, which can highly improve image quality and pleasantness. Furthermore, computer vision algorithms can be more robust to images with reflection, as the reflection can be largely erased by a reflection removal method.

\begin{figure}%[t!]
	\centering
	\begin{tabular}{@{}c@{\hspace{1mm}}c@{\hspace{1mm}}c@{}}
		\includegraphics[width=0.323\linewidth]{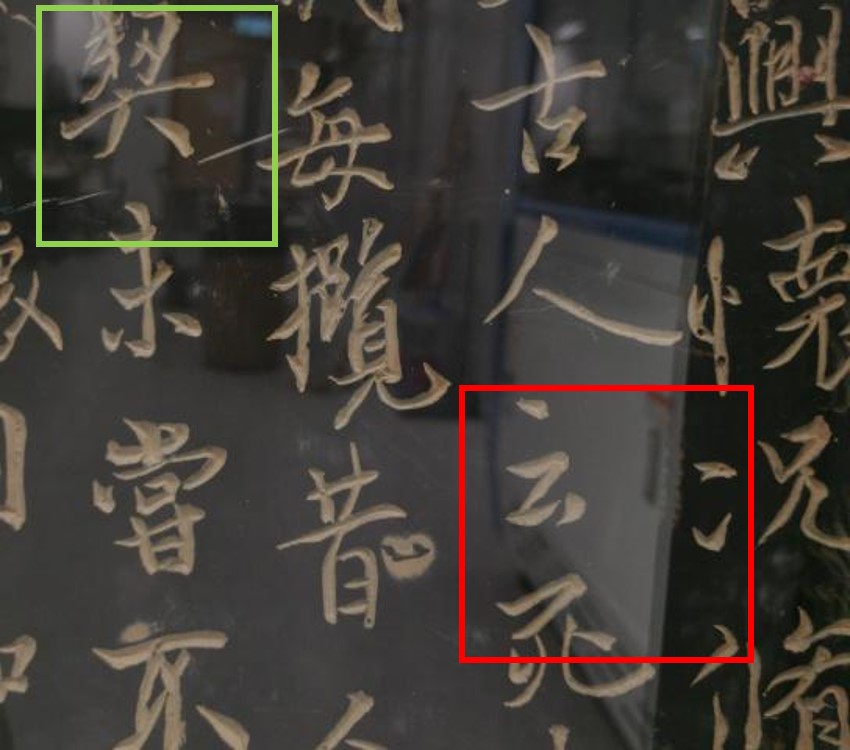} &
		\includegraphics[width=0.323\linewidth]{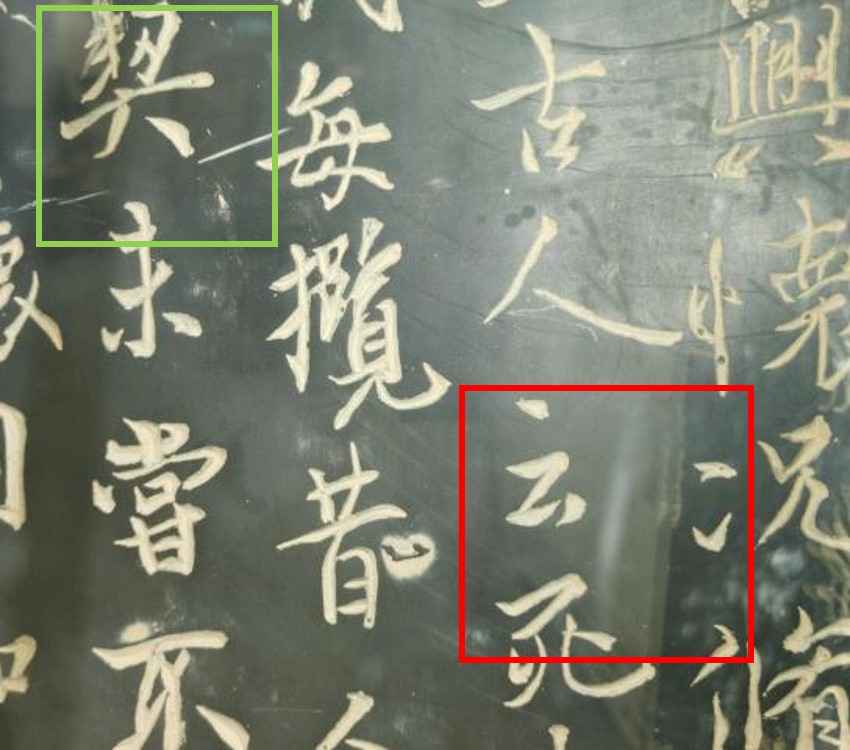}   &
		\includegraphics[width=0.323\linewidth]{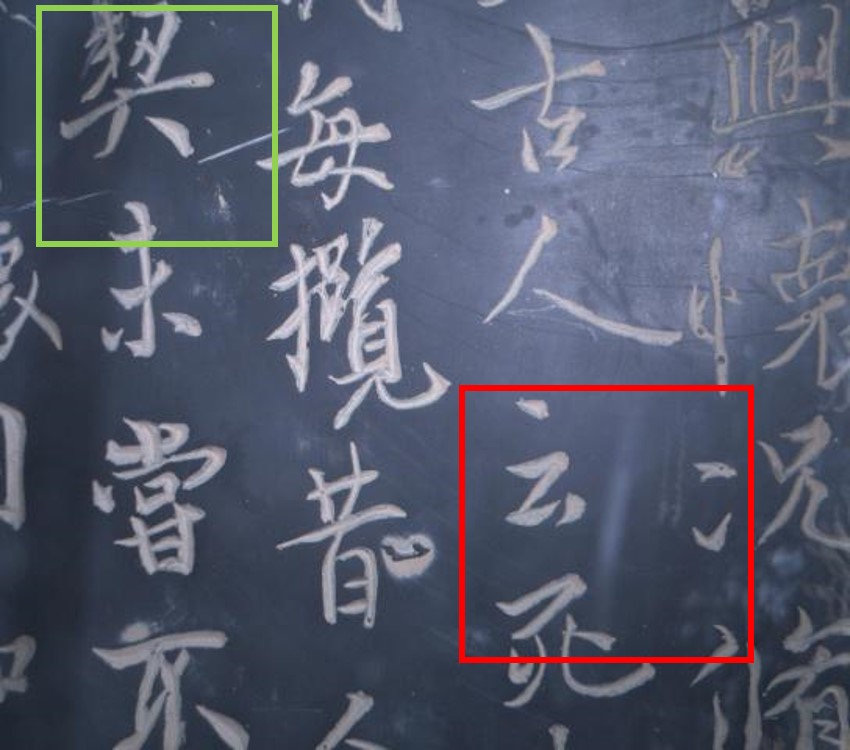}                                \\
		Ambient image                                                                  & Flash image & Flash-only image \\
	\end{tabular}

	\begin{tabular}{@{}c@{\hspace{1mm}}c@{}}
		\includegraphics[width=0.49\linewidth]{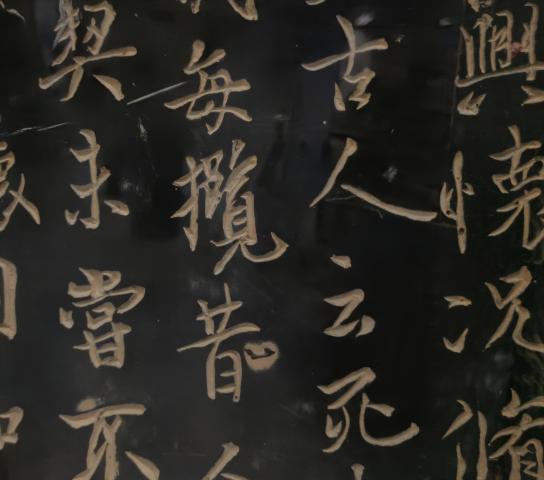} &
		\includegraphics[width=0.49\linewidth]{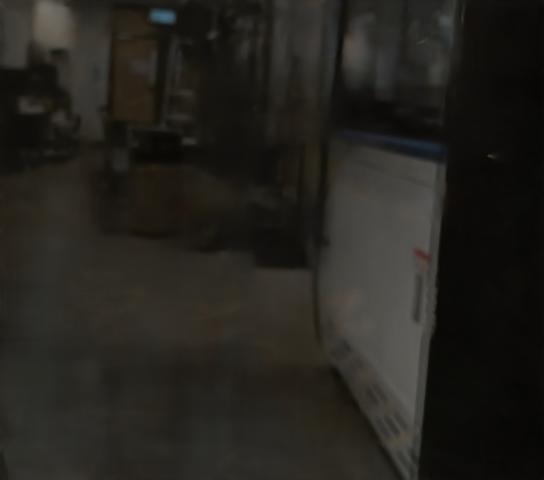}                     \\
		Our transmission                                                                                & Our reflection \\
	\end{tabular}
	\vspace{1mm}
	\caption{A reflection-free flash-only image is computed from a pair of ambient/flash images to help remove reflections. Our transmission image does not absorb the artifacts in the flash-only image.}
	\label{fig:Teaser}
\end{figure}
\begin{figure}%[t!]
	\centering
	\begin{tabular}{@{}c@{\hspace{1mm}}c@{\hspace{1mm}}c@{}}
		\includegraphics[angle=90,width=0.323\linewidth]{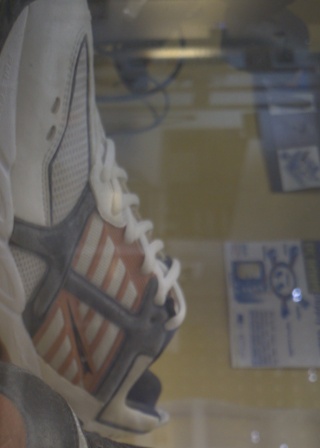} &
		\includegraphics[angle=90,width=0.323\linewidth]{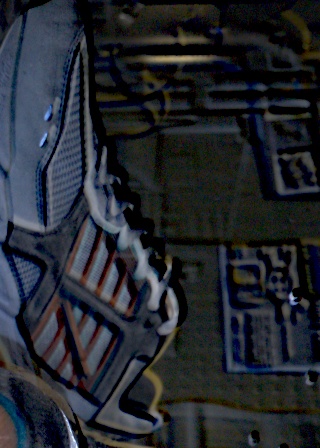}   &
		\includegraphics[angle=90,width=0.323\linewidth]{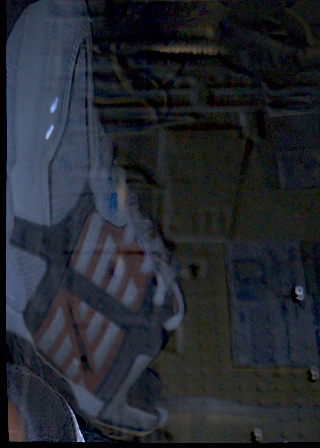}                                \\
		Ambient image                                                                  & Misaligned FO & Aligned FO \\
	\end{tabular}

	\begin{tabular}{@{}c@{\hspace{1mm}}c@{}}
		\includegraphics[angle=90,width=0.49\linewidth]{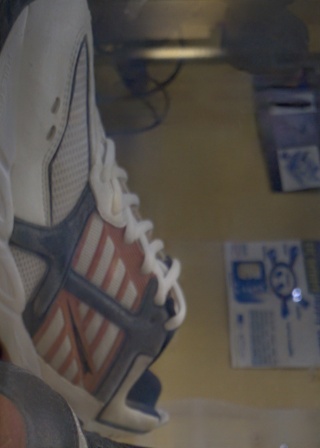} &
		\includegraphics[angle=90,width=0.49\linewidth]{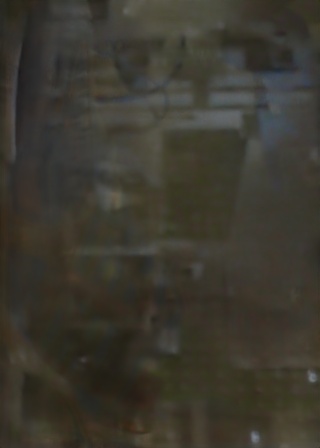}                     \\
		Our transmission                                                                                & Our reflection \\
	\end{tabular}
	\vspace{1mm}
	\caption{For handheld photography, the flash image is first aligned to the ambient image in the transmission layer before the calculation of flash-only. FO: flash-only image.}
	\label{fig:AlignTeaser}
\end{figure}
Reflection removal is challenging because the reflection component is usually unknown. Since both reflection and transmission are natural images, it is hard to distinguish between reflection and transmission from an input image. Therefore, many methods adopt various assumptions on the appearance of reflection for reflection removal. For example, some single-image-based methods~\cite{Arvanitopoulos_2017_CVPR, Yang_2019_CVPR} assume that the reflection is not in-focus and blurry. The ghosting cue~\cite{shih2015reflection_ghost} is another assumption that holds when the glass is thick. However, reflection in real-world images is diverse, and these assumptions do not necessarily hold~\cite{Lei_2020_CVPR,wan2017benchmarking}. As a result, existing methods cannot perfectly remove reflections with diverse appearances from real-world images~\cite{Lei_2020_CVPR}.

We propose a novel \emph{reflection-free flash-only cue} that facilitates inferring the reflection in an ambient image. This cue is robust since it is independent of the appearance and strength of reflection, unlike assumptions adopted in previous single-image reflection removal methods~\cite{shih2015reflection_ghost, Yang_2019_CVPR} or flash-based methods~\cite{chang2020siamese}. The reflection-free cue is based on a physics-based phenomenon of an image obtained by subtracting an ambient image from the corresponding flash image (in raw data space). This flash-only image is equivalent to an image captured under flash-only illumination: the environment is completely dark, and a single flash is the sole light source. A key observation is that the reflection is invisible in the flash-only image.

While flash-only images provide reflection-free cues to distinguish reflection, they also have weaknesses. For instance, in Figure~\ref{fig:Teaser}, we can observe artifacts (e.g., color distortion, illuminated dust) due to uneven flash illumination, occlusions, and other reasons. These artifacts prevent us from obtaining a high-quality transmission easily.

To utilize the reflection-free cue and avoid introducing the flash-only image artifacts, we design a dedicated architecture for obtaining high-quality transmission. Specifically, we first estimate a reflection image instead of a transmission image. Then, to further avoid introducing artifacts in the flash-only image, only the input ambient image and the estimated reflection are given to the second network that estimates the transmission.

Combining our dedicated architecture with the reflection-free cue, we can robustly and accurately remove various kinds of reflection to recover the underlying transmission image. Although we need an extra flash image compared with single-image methods, a flash/no-flash image pair can be captured with a single shutter-press using customized software, as shown in Figure~\ref{fig:datacapcompare}. Hence, general users can easily apply our method for robust reflection removal. 

In summary, our contributions in the preliminary version of this work~\cite{Lei_2021_RFC} are as follows:
\begin{itemize}
	\item We propose a novel cue\textemdash the reflection-free flash-only cue that makes distinguishing reflections simpler for reflection removal. This cue is robust since it is independent of the appearance and strength of reflection.

	\item We propose a dedicated framework that can avoid introducing artifacts of flash-only images while utilizing reflection-free cues. We improve more than 5.23dB in PSNR, 0.04 in SSIM, and 0.068 in LPIPS on a real-world dataset compared with state-of-the-art methods.

	\item We construct the first dataset that contains both raw data and RGB data for flash-based reflection removal.

\end{itemize}

\textcolor{black}{
In this paper, we improve upon our previous work~\cite{Lei_2021_RFC} by extending the capture setting to handheld photography. Specifically, our previous version required a pair of perfectly aligned flash/no-flash images. However, in real-world scenarios, it may be impractical for users to always use a tripod. As a result, misalignments can occur between the flash/no-flash images due to handshake, as shown in Figure~\ref{fig:AlignTeaser}.}

\textcolor{black}{Training on misaligned data is important to achieve robust performance, but collecting a large-scale real-world misaligned flash reflection removal dataset is very challenging. Hence, we propose a misaligned data synthesis pipeline to get abundant training data. A naive way to simulate misalignment is using homography to distort the images, but it cannot model the complicated misalignment: the misalignment between the two images is highly related to depth. For a specific camera motion, pixels with a small distance have large displacements and pixels with large depth can have small displacements. To generate realistic misaligned training data, we use a pretrained depth estimation network to obtain a pseudo depth for each image, which is then used to project the original image according to the depth and camera rotation, and translation.}

\textcolor{black}{To further enhance the performance on handheld captured data, we introduce a dedicated alignment module that aligns the flash/no-flash images. Specifically, we employ the PWC-Net to estimate the correspondence between the flash and no-flash images. We finetune the PWC-Net with our reflection removal network jointly to enhance its robustness to the reflection. To evaluate the real-world performance, we construct a real-world dataset that includes misaligned data for flash-based reflection removal. Our experiments show that our misaligned data and the alignment module improve the performance significantly.}

\textcolor{black}{Finally, we conduct a comprehensive study of the relationship between the linearity of image space and the reflection-free flash-only cue. In our previous work~\cite{Lei_2021_RFC}, our key contribution reflection-free flash-only cue is based on the linearity property of raw images: we manually compute a flash-only image on raw space, which utilizes extra information than a pair of sRGB flash/no-flash images. However, it remains an open question whether it is necessary to compute such a flash-only image on linear image space. Positive results may encourage people to study sophisticated properties of linear RGB images or raw images in different tasks.
}

\section{Related Work}
\subsection{Reflection Removal}

\textbf{Single image reflection removal.}
In single-image reflection removal~\cite{wan2017benchmarking,wan2022benchmarking}, the defocused reflection assumption and ghosting cue are commonly used. The defocused reflection assumption means that reflections are not in focus. Hence, prior work can assume they are more blurry compared with the transmission. Following this assumption, learning-based methods~\cite{fan2017generic,zhang2018single} can synthesize abundant data for training, and non-learning-based methods can suppress the reflection based on image gradient~\cite{Arvanitopoulos_2017_CVPR, Yang_2019_CVPR}. The ghosting cue means multiple reflections are visible on the glass~\cite{shih2015reflection_ghost}. However, the ghosting cue only exists when the glass is thick. Hence, algorithms that are based on the ghosting cue might fail on the thin glass.

There are many attempts to relax assumptions of reflection. Wei et al.~\cite{Wen_2019_CVPR_Linear}
and Ma et al.~\cite{Ma_2019_ICCV} use generative adversarial networks~\cite{DBLP:conf/nips/GoodfellowGAN14} to synthesize realistic reflection under the guidance of real-world reflections. Kim et al.~\cite{Kim_2020_CVPR} propose a physics-based method to render the reflection and mixed image, which improves the quality of training data a lot. Also, Zhang et al.~\cite{zhang2018single}, Wei et al.~\cite{wei2019single_ERR}, Li et al.~\cite{Li_2020_CVPR}, and Lei et al.~\cite{lei2021categorized} collected real-world data to improve the quality of training data. However, as reported by Lei et al.~\cite{Lei_2020_CVPR}, these methods~\cite{Arvanitopoulos_2017_CVPR,fan2017generic, Yang_2019_CVPR} are still far
from perfectly removing reflections for diverse real-world data.

\textbf{Multiple images reflection removal.}
Some reflection removal methods utilize the motion cue of reflection and transmission in multiple images for reflection removal~\cite{guo2014robust,han2017reflection,li2013exploiting,liu2020learning,DBLP:conf/mm/Sun16RR,DBLP:journals/tog/xue2015computational,dualviewrr}. In these motion-based methods, SIFT-flow~\cite{li2013exploiting}, homography~\cite{guo2014robust}, and optical flow~\cite{liu2020learning,DBLP:journals/tog/xue2015computational,dualviewrr} are used to find correspondences among multiple images to distinguish reflection and transmission. However, taking images with different motion cost more effort, and some assumptions are required (e.g., all pixels in transmission must appear in at least one image~\cite{DBLP:journals/tog/xue2015computational}). Polarization is also used in reflection removal to achieve great performance~\cite{Fraid1999,kong14pami,lyu2019reflection,nayar1997separation,eccv2018/Wieschollek, Schechner1999PolarizationbasedDO}. The inputs are usually images through various polarizers, which contain polarization information of light. Since the polarization of reflection and transmission is usually different, it can be used to distinguish them. However, a polarizer is usually required to be shifted to take images, which is complicated. Recently, a camera~\cite{Li_2020_CVPR,Li_eccv20_PolarRR} that can take several polarization images appears, but this kind of camera is yet to be widely used.

\begin{figure*}[t]
	\centering
	\begin{tabular}{@{}c@{}}
		\includegraphics[width=1.0\linewidth]{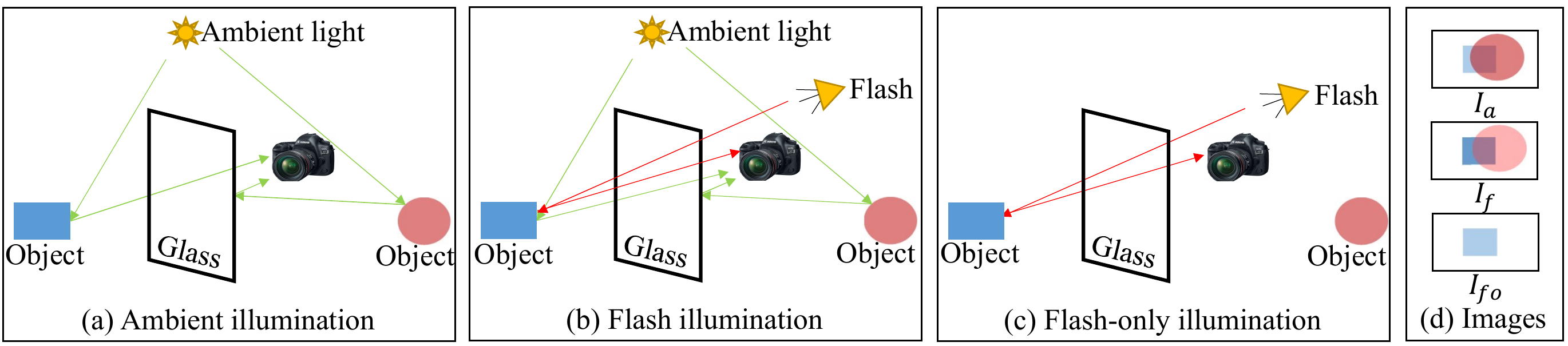} \\
	\end{tabular}
	\caption{An illustration model of the reflection-free cue. Since objects in reflection cannot \emph{directly} receive flash and reflected flash from glass is often weak, flash-only images are visually reflection-free. Note that the flash-only image $I_{fo}$ is obtained from $I_a$ and $I_f$.}
	\label{fig:Illustration.}
\end{figure*}

\begin{figure}[t]
	\centering
	\begin{tabular}{@{}c@{\hspace{1mm}}c@{\hspace{1mm}}c@{\hspace{1mm}}c@{}}
		\rotatebox{90}{ \hspace{2mm}  Sample 1}                                            &
		\includegraphics[width=0.31\linewidth]{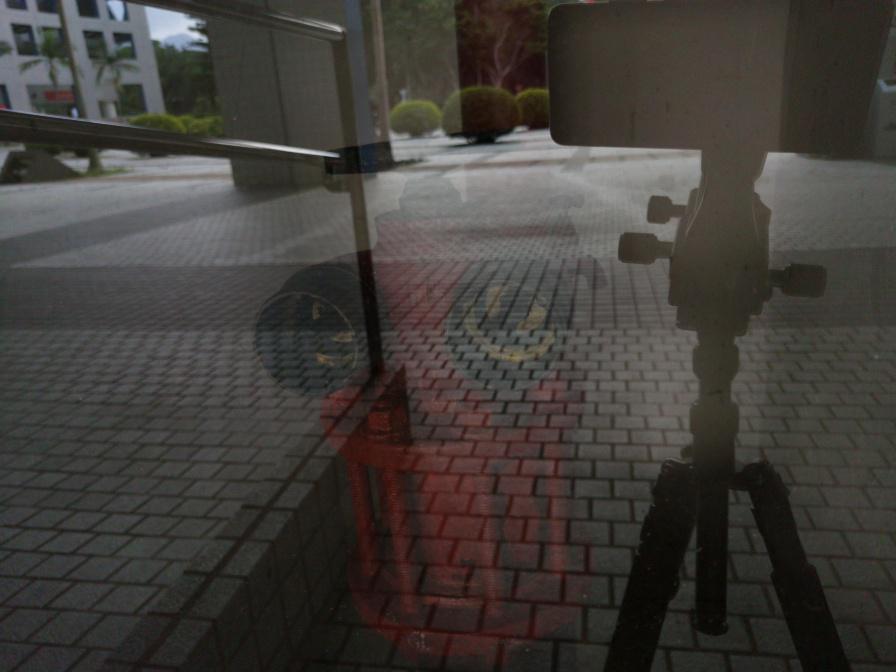} &
		\includegraphics[width=0.31\linewidth]{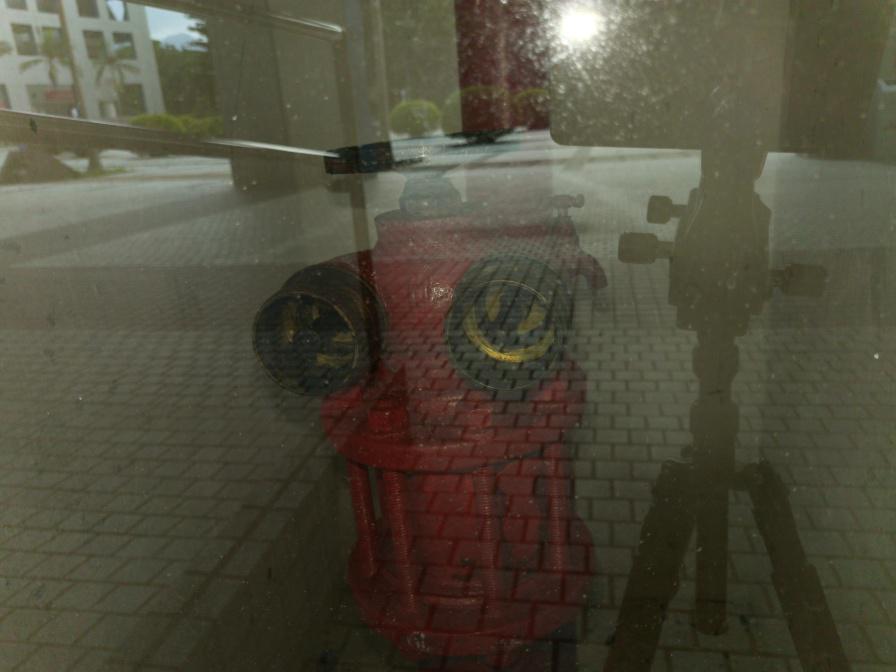}   &
		\includegraphics[width=0.31\linewidth]{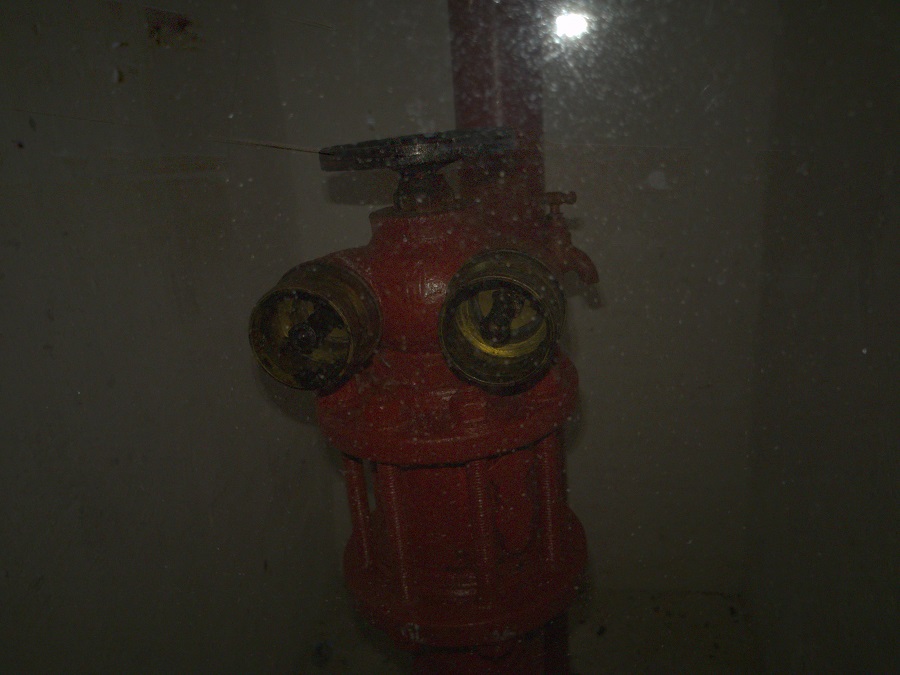}                                                            \\

		\rotatebox{90}{ \hspace{2mm} Sample 2 }                                            &
		\includegraphics[width=0.31\linewidth]{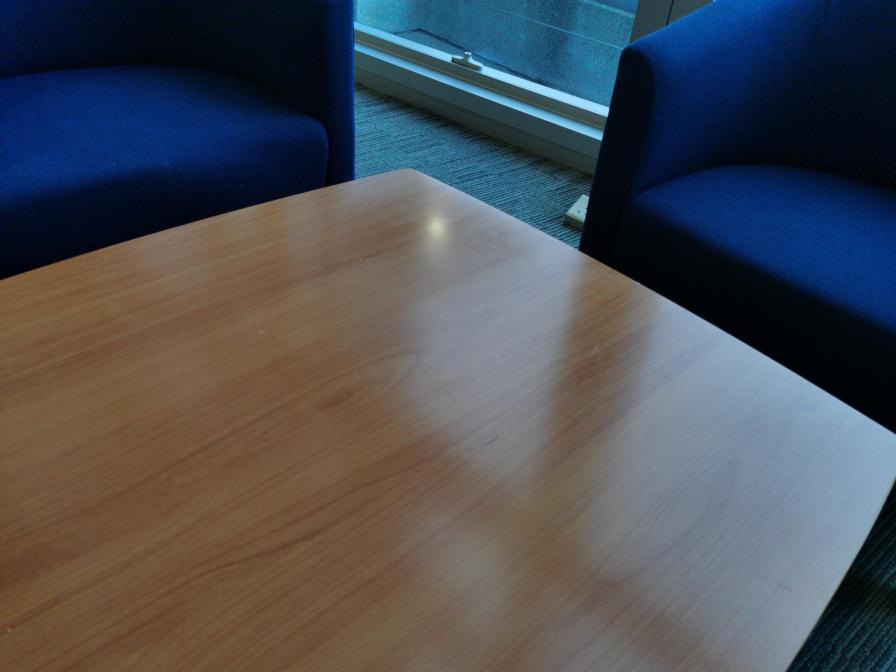} &
		\includegraphics[width=0.31\linewidth]{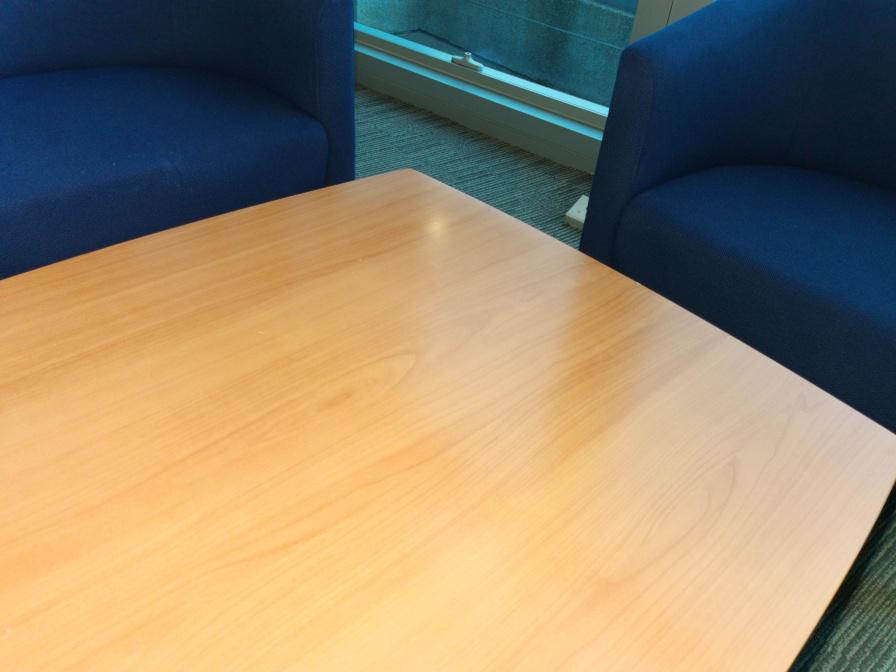}   &
		\includegraphics[width=0.31\linewidth]{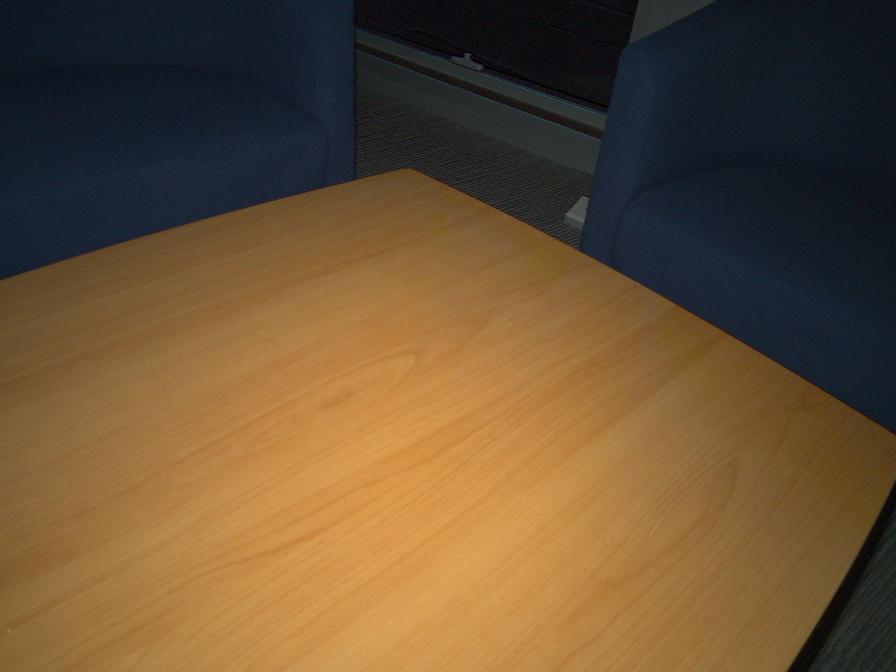}                                                   \\

		                                                                                     & Ambient images & Flash images & Flash-only images \\
	\end{tabular}
	\caption{Examples of reflection-free flash-only images. Flash-only images are visually reflection-free but have many artifacts.}
	\label{fig:pureflash}
\end{figure}

\textbf{Flash-based reflection removal.}
Various properties of a pair of flash/ambient images are adopted in previous work~\cite{agrawal2005removing_flash, chang2020siamese}. Agrawal et al.~\cite{agrawal2005removing_flash} claim that gradient orientations are consistent in the image pair, assuming that depth edges, shadows, and highlights are few.
However, they cannot generate reasonable results for undesirable regions (e.g., shadows, specular reflection), and their results tend to be over-smooth. SDN~\cite{chang2020siamese} utilizes the assumption that reflection can be obviously suppressed by flash, but the suppression effect is sensitive to the strength of reflection: when reflection is strong, the suppression is no longer effective.

\subsection{Flash Photography}
A pair of flash and no-flash images captured consecutively is usually used in flash photography. The extra flash image helps obtain novel capability or enhance the performance in various computational photography tasks. Petschnigg et al.~\cite{DBLP:journals/tog/PetschniggSACHT04} use the flash image for denoising, detail transfer, etc. Drew et al.~\cite{drew2006removing} use the flash-only image for shadow removal. Sun et al.~\cite{sun2005flashmatting,sun2007flashcut} observe that the change of intensity is different for near objects and background in the flash-only image and apply it to image matting~\cite{sun2005flashmatting} and foreground segmentation~\cite{sun2007flashcut}. Using the property that there is only a single light source under the flash-only image, Hui et al. propose a closed-form white balance algorithm~\cite{hui2016white}. Hui et al.~\cite{hui2018illuminant} propose an approach to separate light sources under complicated illumination. Cao et al.~\cite{cao2020stereoscopic} use the flash-only image for shape and albedo recovery under the assumption of the Lambertian model. Compared to prior works, we notice a novel property lies under flash-only images for reflection removal.

\textcolor{black}{Hong et al.~\cite{hong2020near,hong2022reflection} have proposed to use NIR images captured by an active NIR camera for reflection removal. Specifically, they use a projector to emit the NIR light and utilize the observation that the captured NIR images are less likely to be affected by reflections for certain angles between glass and camera planes. We share the similarity that both Hong et al.~\cite{hong2020near,hong2022reflection} and our method require active illumination. However, our approach does not require an active NIR projector and the corresponding NIR camera. Instead, we use the ubiquitous flashlight that is present on most mobile phones and cameras. Besides, reflections are only weak for certain viewing angles in NIR images, while our flash-only images do not contain reflections of ambient images for different viewing angles in principle. 
}

\begin{figure*}%[t]
	\centering
	\begin{tabular}{@{}c@{}}
		\includegraphics[width=1.0\linewidth]{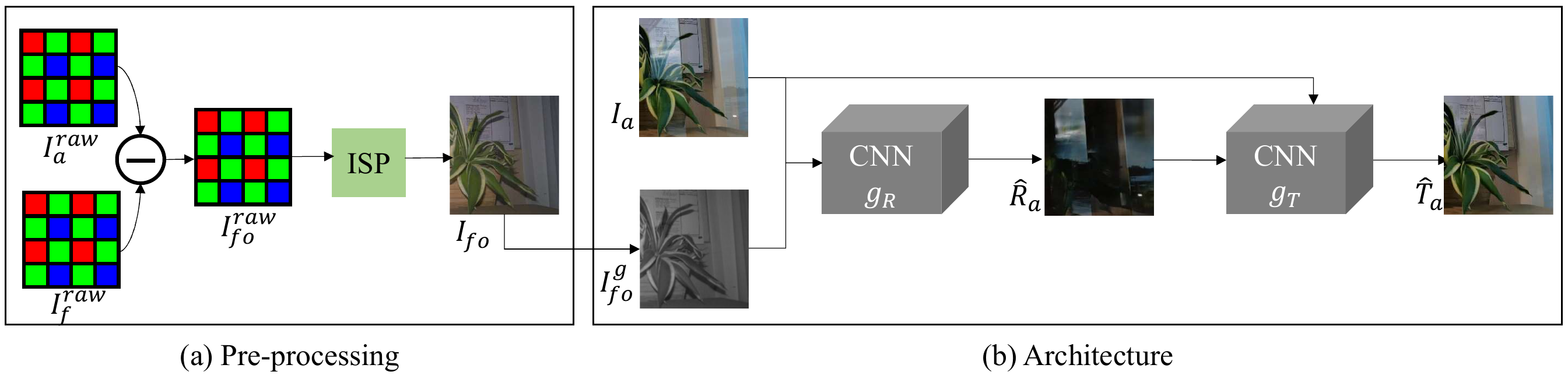}
	\end{tabular}
	\caption{The overall architecture of our approach. We compute the $I_{fo}$ from $I_f^{raw}$ and $I_a^{raw}$. Then, our dedicated architecture estimates the reflection first to avoid absorbing artifacts of flash-only images. Finally, the transmission is estimated with the guidance of reflection. }
	\label{fig:Architecture}
\end{figure*}

\section{Reflection-free Flash-only Cues}
\label{sec:flashonly}

\textbf{Flash-only images.}
Let $\{I_a^{rgb}, I_{f}^{rgb}\}$/$\{I_a^{raw}, I_{f}^{raw}\}$ be the RGB/raw images under ambient and flash illuminations. A flash-only image $I_{fo}^{raw}$ can be computed from $I_a^{raw}$ and $I_{f}^{raw}$. Since the flash image is the sum of the ambient image and flash-only image for a linear response camera in linear space~\cite{DBLP:journals/tog/PetschniggSACHT04}, we can obtain the flash-only image through:
\begin{align}
	\label{eq:pureflash}
	I_a^{raw}    & = R_{a}^{raw} + T_a^{raw},                              \\
	I_{f}^{raw}  & =R_{a}^{raw} + T_a^{raw} + R_{fo}^{raw} + T_{fo}^{raw}, \\
	I_{fo}^{raw} & = I_{f}^{raw} - I_a^{raw}=R_{fo}^{raw} + T_{fo}^{raw},
\end{align}
where $R$ and $T$ are the reflection and transmission. For simplicity, we also use $I^{raw}$ to denote the image after linearization. Reflection-free cues exist in flash-only images. The flash-only image is equivalent to an image captured in a completely dark environment, and the flash is the sole light source, as shown in Figure~\ref{fig:Illustration.}(c). Note that $I_{fo}^{raw}$ is invariant to different ambient illuminations as long as $I_a^{raw}$ and $I_f^{raw}$ do not have saturated pixels. 

\textbf{Reflection-free cues.}
The reflection-free cue denotes a physics-based phenomenon: reflections of the ambient image $R_a^{raw}$ are invisible in the flash-only image. Besides, we have $R_{fo}^{raw}\approx \mathbf{0}$ in most cases. Figure~\ref{fig:Illustration.} shows an illustration of this phenomenon. Reflections exist in ambient images because objects in the reflection receive ambient light and then reflect it to the camera through the glass. In Figure~\ref{fig:Illustration.}(b), objects in reflection do not directly receive light from the flash. \textcolor{black}{It can only receive the 
reflected flashlight from the glass. The objects in the reflection then reflect the light back onto the glass, and finally the glass reflects the light towards the camera. That is to say, the intensity of the light is reduced twice when it is reflected by a glass surface. Besides, the reflectance of glass is mostly much weaker than transmittance. Hence, objects in reflection are barely illuminated and almost invisible in the flash-only images. }

To verify reflection-free cues, we capture pairs of ambient and flash images under different illumination and scenes, and we compute $I_{fo}^{raw}$ following Eq.~\ref{eq:pureflash}. As shown in Figure~\ref{fig:pureflash}, reflections disappear in flash-only images, even when reflections are strong. Also, the second example shows this cue is valid not only for semi-reflecting surfaces.

\textbf{Undesirable artifacts.}
Although flash-only images are visually reflection-free, they usually have undesirable artifacts, as shown in Figure~\ref{fig:pureflash}. We can analyze the reasons of degradation formally from {flash-only radiance} by Eq.~\ref{eq:radiance}~\cite{kajiya1986rendering}:
\begin{align}
	\label{eq:radiance}
	L_o^{fo}          & = \int_{\Omega}f_r({\omega}_i,\omega_o)L_i(\omega_i)(\omega_i \cdot \mathbf{n})d \omega_i, \\
	\label{eq:Falloff}
	L_{i,d}^{fo}(w_i) & = \frac{L^{fo}(w_i)}{d^2},
	% \\
	% \label{eq:BRDF}
	% f_r(\omega_i^{fo},\omega_o) &=\frac{f_r^d}{\pi}+f_r^s(\omega_i^{fo},\omega_o),
\end{align}
where $L_o$ and $L_i$ are the radiance of outgoing and incident light, $\omega_i$ and $\omega_o$ are the light direction of outgoing and incident light, $f_r$ is the bidirectional reflectance distribution function (BRDF), $\mathbf{n}$ is the surface normal, and $\Omega$ is the hemisphere. Flash-only images can contain the following artifacts that require to be resolved:

(1) Color distortion usually appears since the flash light $L^{fo}$ is different from ambient illumination. Similarly, the shading also changes since the light direction $w_i$ is different.

(2) Uneven illumination is a common problem due to irradiance falloff in Eq.~\ref{eq:Falloff}, irradiance (and thus radiance $L_{i,d}^{fo}$) is different due to different distance $d$ to the flash.

(3) New shadows are brought by occlusion.

(4) Highlights caused by flash might appear on the glass.

(5) If the glass is dirty, dust can be illuminated on the glass, as shown in 1st example in Figure~\ref{fig:pureflash}.

\section{Method}

Given an ambient image $I_a^{raw}$ and a flash image $I_f^{raw}$ in raw space, our approach aims to estimate the transmission $T_a$ under ambient illumination. In Figure~\ref{fig:Architecture}(a), we first take raw images $I_a^{raw}, I_f^{raw}$ and implement pre-processing to obtain the RGB flash-only image $I_{fo}$, as introduced in Section~\ref{sec:data_prepare}. Then, our dedicated architecture in Section~\ref{sec:Architecture} takes RGB images $I_a, I_{fo}$ as input to remove the reflection.

\subsection{Pre-processing}
\label{sec:data_prepare}
In this section, we prepare the images that are used for training the neural networks in Section~\ref{sec:Architecture}.

We first capture two raw images $I_a^{raw}$ and $I_f^{raw}$ through the pipeline in Section~\ref{sec:Dataset}. Given $I_a^{raw}$ and $I_f^{raw}$, we implement the following pipeline to obtain RGB images:

1) Subtraction. We first implement linearization to convert images to linear space using the black-level and white-level information from the metadata. After this step, the range of each pixel is transferred to $[0, 1]$. Then the flash-only image is obtained through Eq.~\ref{eq:pureflash} since the linearity between pixel values and physical light is preserved well. At last, the flash-only image is converted back to raw space using the black-level and white-level information.

2) Image signal processing (ISP). We implement a regular ISP~\cite{2016ISP} that includes linearization, demosaiced, white balance, color correction, and gamma correction to convert raw images to RGB images using the original metadata of images. We adopt the metadata of $I_a^{raw}$ to process $I_{fo}^{raw}$ since it is obtained by $I_a^{raw}, I_f^{raw}$, and no metadata is available. Note that the white balance of $I_{fo}$ is usually not as good as $I_a$ since it does not have its own metadata. For $I_a$ at test time, we can use our ISP to obtain the sRGB image or use the original sRGB image processed by the camera's ISP. A learning-based ISP~\cite{Chen2018SID,xing2021cvpr,Zamir2020CycleISP} can also be used here.

\subsection{Architecture}
\label{sec:Architecture}

We first describe a naive model for reflection removal, which introduces undesirable artifacts in the estimated transmission. To handle the problem of the naive model, we propose a two-stage anti-artifacts architecture that blocks the artifacts in the first stage.

\subsubsection{Flawed naive model}

As shown in Figure~\ref{fig:pureflash}, the reflection in $I_a$ does not exist in $I_{fo}$. Except for the artifacts and color distortion, the $I_{fo}$ is quite similar to our target transmission $T_a$. Hence, we first try to use a network to directly estimate transmission from $I_a$ and $I_{fo}$, which we denote as base model $g_B$ (note that this model $g_B$ is \emph{not our final model}):
\begin{align}
	\hat T_B = g_B(I_a,I_{fo};
	\theta_B),
\end{align}
where $\theta_B$ is the parameters of network $g_B$. However, we observe that although this model can correctly remove various types of reflections$R_a$, the estimated $\hat T_B$ has undesirable artifacts, especially in the area that contains shadows, highlight in flash-only images. Also, the color might be closer to $I_{fo}$ in some areas (i.e., color distortion), as shown in Figure~\ref{fig:AblationStudy}(d).

We argue that: since the transmission component is the intersection of $I_{fo}$ and $I_a$, inputting the flash-only image $I_{fo}$ to the transmission-estimation network can introduce artifacts in the final prediction.

\subsubsection{Reflection-pass network}
To avoid inputting the flash-only image $I_{fo}$ to the transmission-estimation network and leverage the reflection-free property of flash-only images, we thus design a two-stage network in Figure~\ref{fig:Architecture}: we only estimate the reflection first instead of directly estimating $T_a$ and then use the estimated reflection to predict transmission $T_a$. Stated in another way, only the reflection passes the first network. As reflections $R_a$ only exist in $I_a$, it must be extracted from $I_a$ and avoid introducing artifacts of $I_{fo}$. On the other hand, since there is no reflection in the flash-only image $I_{fo}$, it can provide strong guidance for reflection estimation. Specifically, we first convert the flash-only image to a grayscale image $I_{fo}^g$ to avoid the influence of color distortion. In practice, we find that the grayscale flash-only image can provide enough structure information for estimating the reflection. Then, $I_a, I_{fo}^g$ are concatenated as input to the network $g_R$:
\begin{align}
	\hat R_a           & = g_R(I_a,I_{fo}^g; \theta_R), \\
	L_R(R_a, \hat R_a) & = ||R_a - \hat R_a||_2^2,
\end{align}
where $\theta_R$ is the parameters of network $g_R$. We adopt the L2 loss for training $g_R$.

\begin{figure}[t!]
	\centering
\begin{tabular}{@{}c@{\hspace{1mm}}c@{\hspace{1mm}}c@{\hspace{1mm}}c@{}}
\rotatebox{90}{\hspace{4mm} \small Input}&
\includegraphics[width=0.305\linewidth]{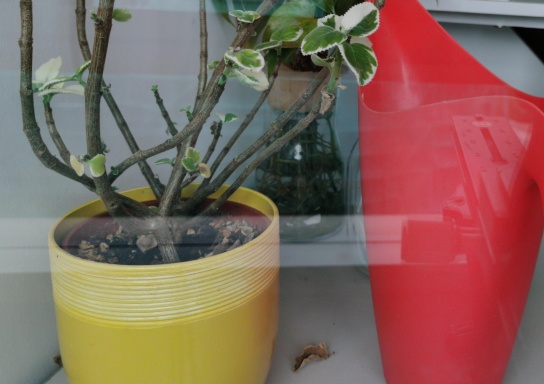}&
\includegraphics[width=0.305\linewidth]{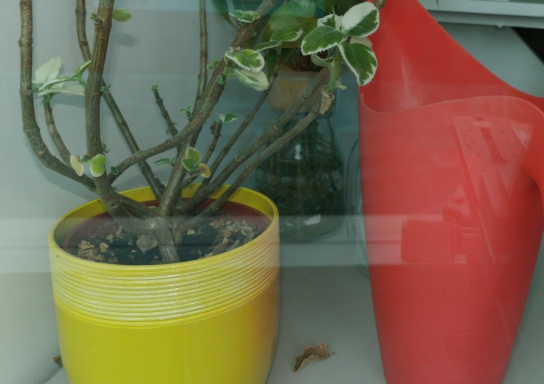}&
\includegraphics[width=0.305\linewidth]{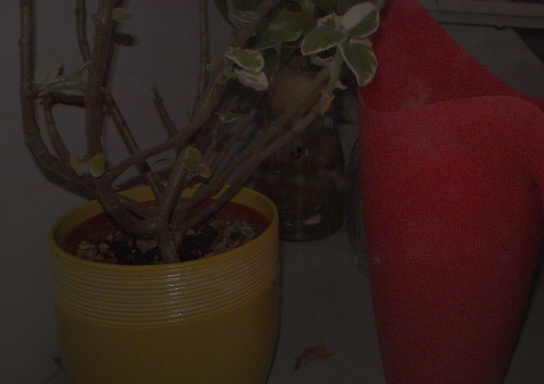}\\
&(a) Input $I_a$ &(b) Input $I_f$ & (c) Processed $I_{fo}$\\
\rotatebox{90}{\hspace{4mm} \small Our $\hat T$}&
\includegraphics[width=0.305\linewidth]{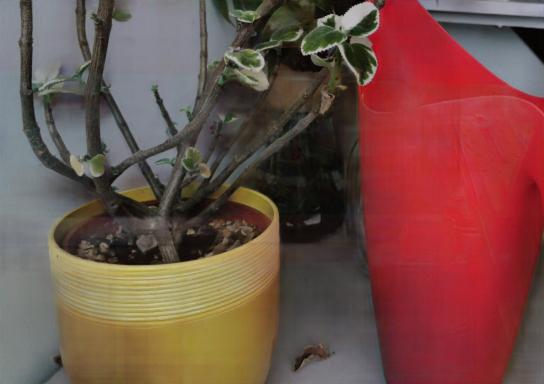}&
\includegraphics[width=0.305\linewidth]{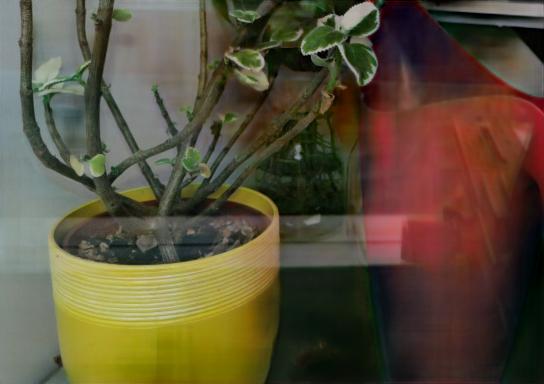}&
\includegraphics[width=0.305\linewidth]{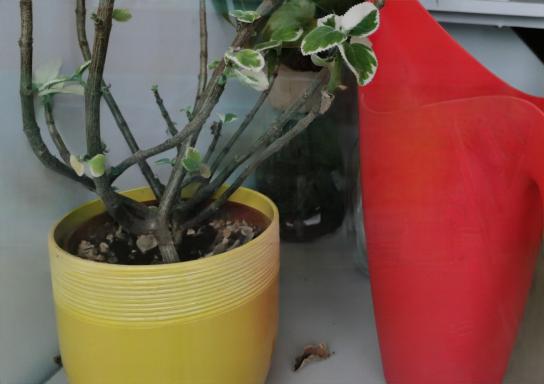}\\
&{(d) $g_B$, $I_{fo}$}   & (e) {$g_R+g_T$, $I_f$ }&  
{(f) $g_R+g_T$, $I_{fo}$}\\
\end{tabular}

	\vspace{1mm}
	\caption{Qualitative comparison among multiple implementations. Combining the reflection-free $I_{fo}$ with our dedicated architecture achieves the best performance.}

	\label{fig:AblationStudy}
\end{figure}
The effectiveness of adopting reflection as guidance has been proven in previous work~\cite{Lei_2020_CVPR, Li_2020_CVPR,eccv18refrmv_BDN}. Hence, we directly use the estimated reflection and the ambient image to estimate the transmission. Note that the flash-only image $I_{fo}$ is \emph{not input} to $g_T$ to avoid introducing artifacts. The transmission $\hat{T_a}$ is then estimated:
\begin{align}
	\hat T_a & = g_T(I_a, \hat R_a; \theta_T),
\end{align}
where $\theta_T$ is the parameters of $g_T$. We also adopt a L2 loss for training $g_T$. As shown in Figure~\ref{fig:AblationStudy}(f), the result of $g_R + g_T$ does not contain obvious artifacts (e.g., color distortion), which is much better than the result of $g_B$ in Figure~\ref{fig:AblationStudy}(d).

\subsubsection{Training details}
We train for 150 epochs with batch size 1 on an Nvidia RTX 2080 Ti GPU. We use the Adam optimizer~\cite{DBLP:journals/corr/KingmaB14} to update the weights with an initial learning rate of $10^{-4}$ and cosine annealing schedule. The plain U-Net~\cite{Ronneberger2015} is used for the two networks (with trivial modification~\cite{lei2020dvp}). The two networks $g_R$ and $g_T$ are trained simultaneously.
We evaluate the inference time of our proposed method with alignment on an Nvidia RTX 2080 Ti GPU. 
The inference time of a flash/no-flash pair with a resolution $640 \times 640$ is 65ms.

\begin{figure*}[t]
	\centering
	\begin{tabular}{@{}c@{}}	\includegraphics[width=\linewidth]{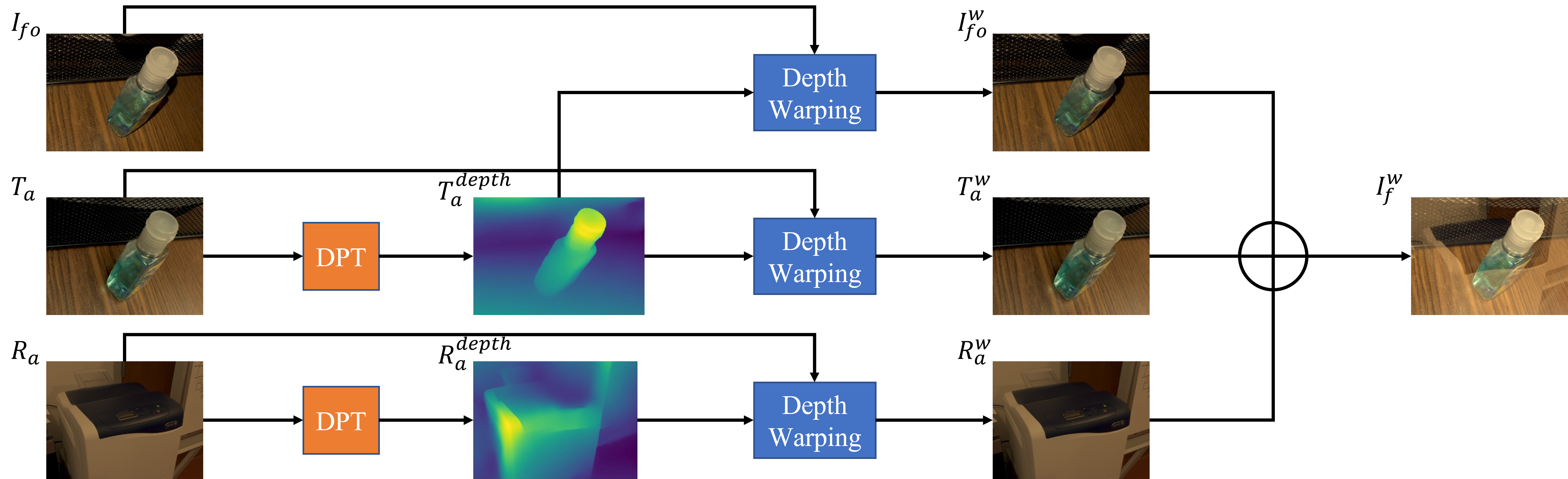}
	\end{tabular}
	\caption{{Misaligned data synthesis}. Depths for transmission and reflection are estimated by a pretrained monocular depth estimation network DPT~\cite{Ranftl2021}. Then warped images are rendered with the same random camera motion. Finally, warped flash-only, ambient transmission, and ambient reflection images are added in linear space to obtain the misaligned flash image. }
	\label{fig:SynMisalign}
\end{figure*}

\begin{figure}[t]
	\centering
	\begin{tabular}{@{}c@{}}
		\includegraphics[width=1\linewidth]{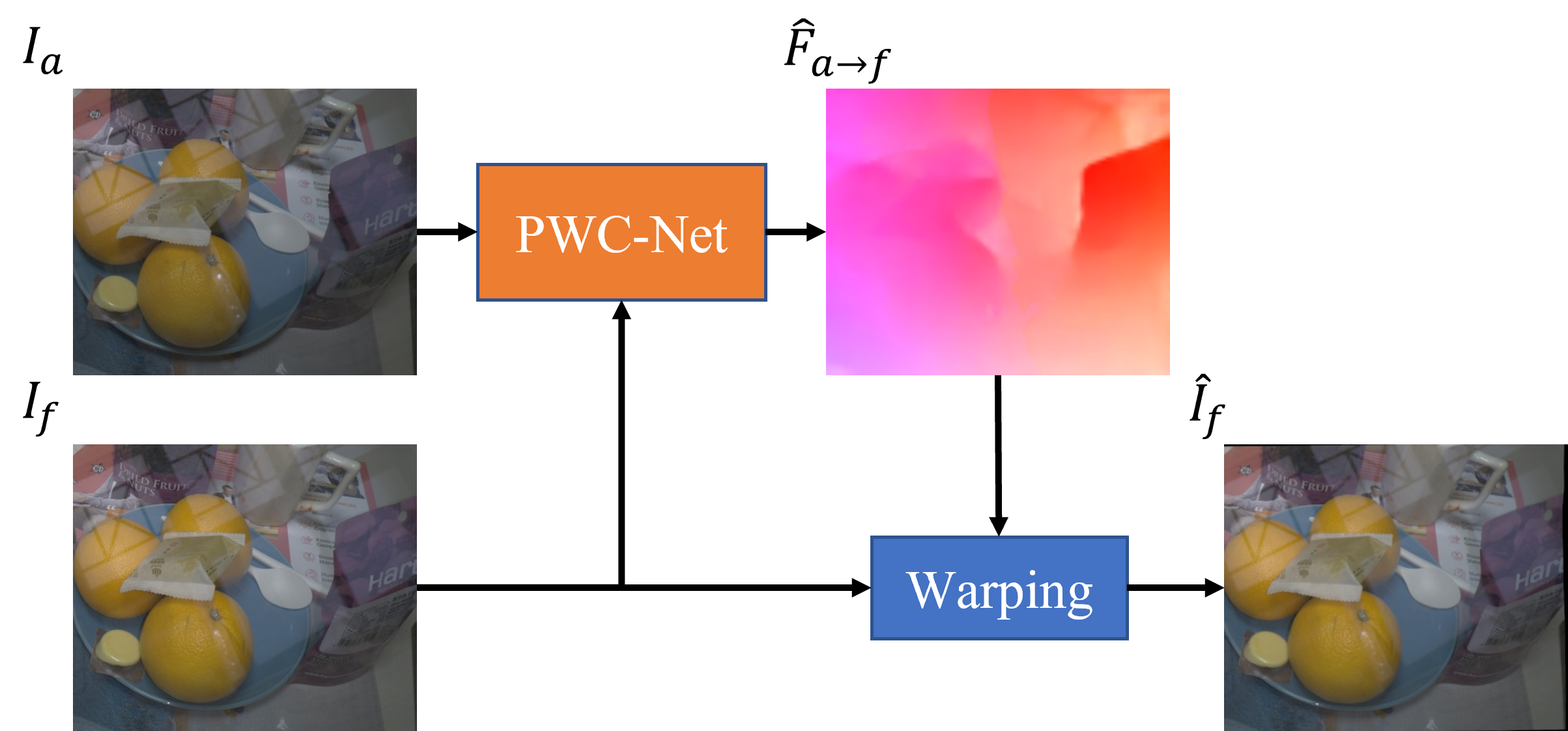}
	\end{tabular}
	\caption{{Alignment module}. To solve the misalignment problem between the flash and ambient images in handheld photography, PWC-Net~\cite{Sun2018:Model:Training:Flow} is applied to estimate the optical flow of the transmission layer. }
	\label{fig:Alignment Module}
\end{figure}

\begin{figure*}
	\centering
	\begin{tabular}{@{}c@{}}
		\includegraphics[width=1.0\linewidth]{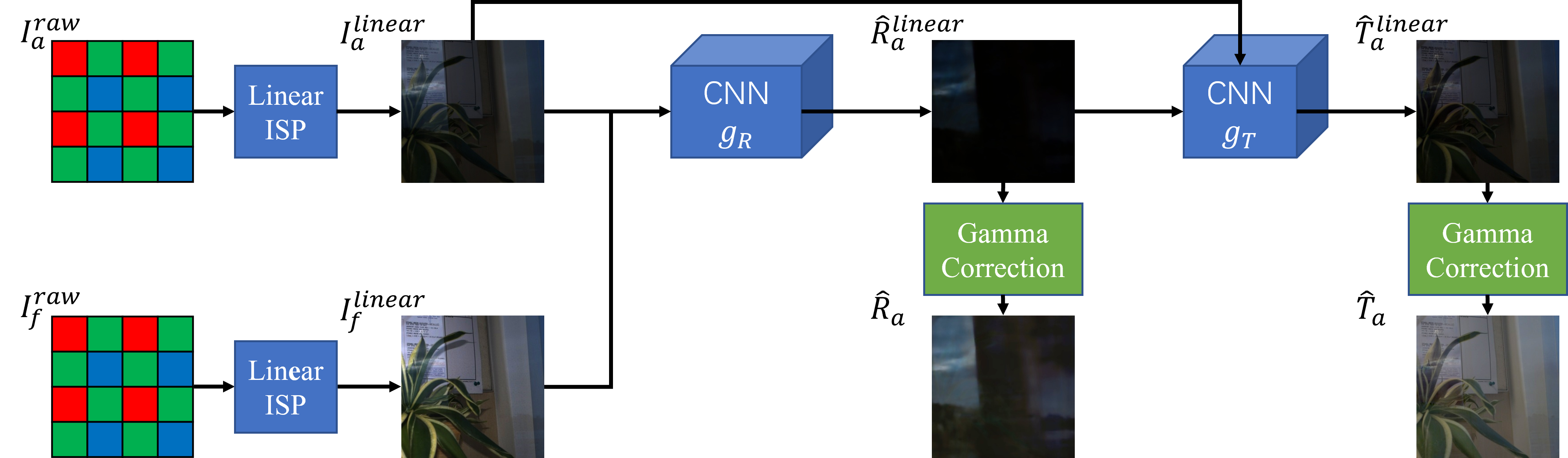}
	\end{tabular}
	\caption{\textcolor{black}{The overall architecture of our approach with linear RGB images as input. In our previous version, we argued that using flash-only images can achieve better performance than using flash images on the RGB image space. However, in this version, we show that using flash images in \textbf{the linear space} can achieve similar performance to using flash-only images. Therefore, we conclude that the key to the success of flash-only images is the linearity property. This observation provides insight into the importance of raw images for reflection removal and other image processing tasks. The linear ISP consists of linear operations, including linearization, demosaicing, and white balance.}}
	\label{fig:ArchitectureLinear}
\end{figure*}

\subsection{Extension to Handheld Misaligned Data}
\label{subsec:misalign}
In this section, we aim to extend our approach to handheld photography since users usually cannot take a tripod easily in daily life. In Section~\ref{sec:data_prepare}, we assume that the flash/no-flash pair is perfectly aligned so that we can obtain a flash-only image. However, in handheld photography, it is hard to capture two images with perfect alignment without a tripod. We propose two key factors that can help our model achieve robust and satisfying performance in handheld photography: (1) proposing a misaligned data synthesis pipeline to train the model, and (2) designing an alignment module to align the flash/no-flash pair.

\subsubsection{Training on misaligned data}
We notice that the performance of our model is degraded a lot when it receives misaligned data at the test time, as reported in the experiment section. The degradation is explainable since all the flash/no-flash pairs are perfectly aligned at the training time, which is quite different from the test setting. To solve this problem, we propose to train the model on misaligned data.

It is time-consuming to collect a large-scale misaligned dataset for training. \textcolor{black}{To simulate misalignment between flash and no-flash images, a common approach is to apply random homography. However, using homography cannot model the complex misalignment that occurs due to depth differences. Hence, inspired by recent progress in optical flow data synthesis~\cite{Aleotti2021stillflow}, we propose a data synthesis pipeline that can generate realistic misaligned training data with a pretrained depth estimation network.} 

Figure~\ref{fig:SynMisalign} shows the pipeline to synthesize misaligned data. We first estimate the depth of the transmission and reflection through a pretrained depth estimation network DPT~\cite{Ranftl2021}. Then, with a random camera motion, we warp the flash-only image $I_{fo}$ and ambient transmission image $T_a$ according to the same estimated transmission depth $T_a^{depth}$. With the same camera motion, we warp the reflection image $R_a$ according to the reflection depth $R_a^{depth}$. At last, the warped flash-only, transmission, and reflection images are added in linear space to obtain the misaligned flash image $I_f^w$. In addition, the corresponding ground truth optical flows of the transmission and reflection warping are also available. In Section~\ref{sec:Analysis}, we show that training on misaligned data can efficiently improve performance without changing the framework.

\subsubsection{Alignment for preprocessing}
While training on misaligned data can improve performance, we notice that degradation still exists when the misalignment is large. We propose to align the flash and no-flash pair before the subtraction in Section~\ref{sec:data_prepare}. Figure~\ref{fig:Alignment Module} shows the framework of our alignment module. Specifically, we estimate the transmission layer optical flow from the ambient image to the flash image. Then we warp the flash image $I_f$ to a warped flash image $\hat I_f$ that is aligned with the ambient image in the transmission layer. 

\textcolor{black}{The quality of the estimated flow of the pretrained flow estimation network can be affected by the reflections in images. To address this, we finetune the PWC-Net ~\cite{Sun2018:Model:Training:Flow} using the flow of the transmission layer during data synthesis. This approach enhances the robustness of our method to the presence of reflection in the images. We show that this alignment module can improve the performance on real-world misaligned data in the experiments.}

\subsection{Additional Challenges in the Wild}

\textcolor{black}{We discuss additional challenges in the wild beyond the misalignment inputs in this part.}

\begin{figure}[h]
	\centering
	\begin{tabular}{@{}c@{\hspace{1mm}}c@{\hspace{1mm}}c@{\hspace{1mm}}c@{}}

		\includegraphics[width=0.24\linewidth]{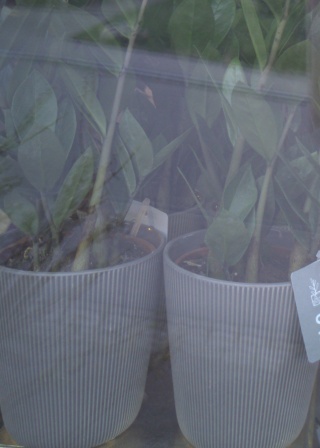}   &
		\includegraphics[width=0.24\linewidth]{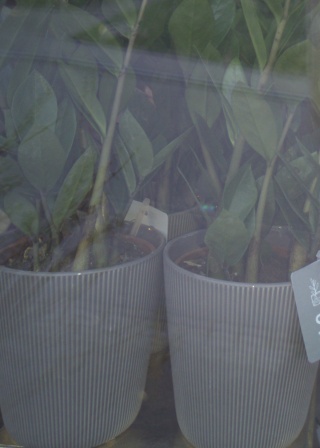}          &
  		\includegraphics[width=0.24\linewidth]{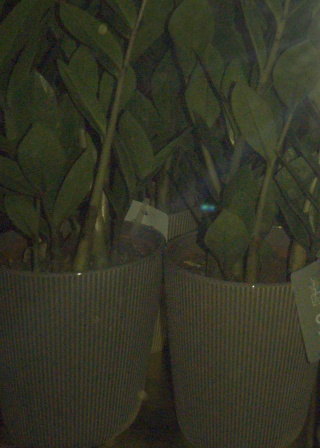}          &
		\includegraphics[width=0.24\linewidth]{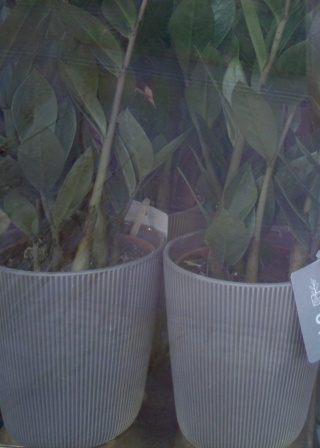} \\

		Input $I_a$                                                                                          &Input  $I_f$                                                                                          & Processed $I_{fo}$ & Ours \\
		(60 ms) & (40 ms)  & &  \\

	\end{tabular}
	\vspace{1mm}
	\caption{Our model can handle different exposure times for the ambient image and the flash image.}
	\label{fig:exposure}
\end{figure}

\textcolor{black}{\textbf{Varied exposure time.} In practical situations, it may be necessary to adjust the exposure of the two images to avoid underexposure or overexposure, which does not follow the assumption of Eq.~\ref{eq:pureflash}. For example, in low-light conditions, the ambient image may require a long exposure time, but using the long exposure time for the flash image could result in overexposure. Hence, we adopt the following approach to handle inconsistent exposure times: }
\begin{align}
	I_{fo}^{raw} & = I_{f}^{raw} - \frac{e_f}{e_a}I_a^{raw},
\end{align}
where $e_a$ and $e_f$ be the exposure time for the ambient image and flash image respectively. The computed flash-only image is then inputted to the neural network to remove the reflection, as shown in Figure~\ref{fig:exposure}.

\begin{figure}[t!]
	\centering
	\begin{tabular}{@{}c@{\hspace{1mm}}c@{\hspace{1mm}}c@{}}

		\includegraphics[width=0.30\linewidth]{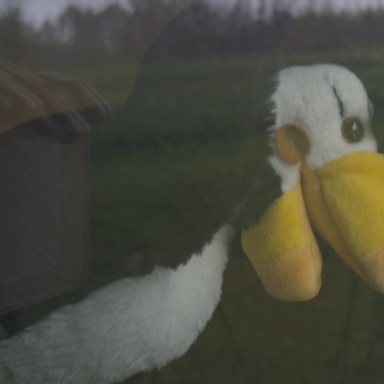} &\includegraphics[width=0.30\linewidth]{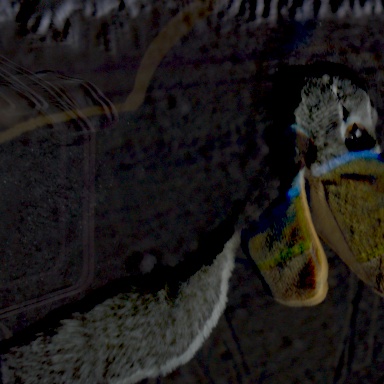} &\includegraphics[width=0.30\linewidth]{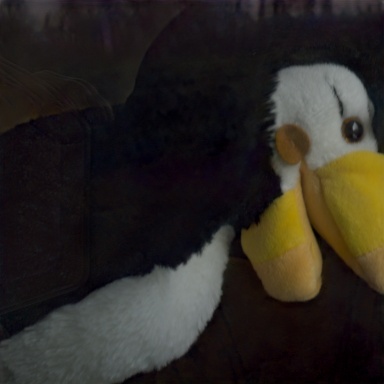} \\
  Input $I_a$   & Before align. $I_{fo}$ & Ours $\hat{T}$ \\
  \includegraphics[width=0.30\linewidth]{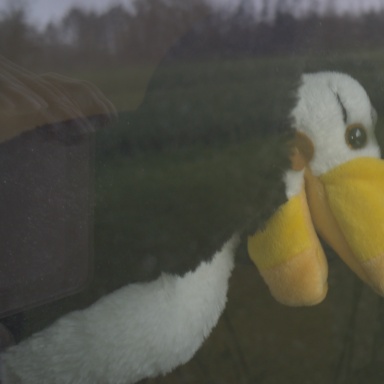} &\includegraphics[width=0.30\linewidth]{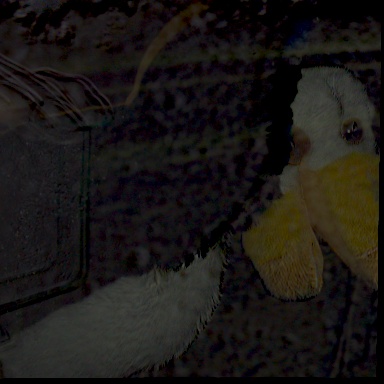} &\includegraphics[width=0.30\linewidth]{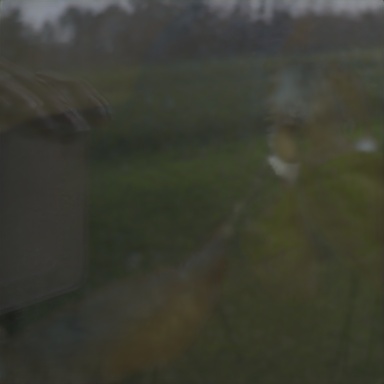} \\
  Input  $I_f$ & After align. $I_{fo}$ & Ours $\hat{R}$
	\end{tabular}
	\vspace{1mm}
	\caption{Qualitative result on an image pair of a hand puppet bird in motion (moving and flapping the wings). Our model can handle moderate motion of transmission and achieve satisfying performance.}
	\label{fig:dynamic}
\end{figure}

\textcolor{black}{\textbf{Dynamic of transmission scenes.} In Eq.~\ref{eq:pureflash}, we assume that the transmission scenes are static in two shots. In practice, the dynamics of transmission scenes can be affected by factors such as object motion or changes in illumination. However, since the time interval between two consecutive shots is usually very short, differences are typically small. Additionally, our model can handle moderate object motion in transmission scenes, as shown in Figure~\ref{fig:dynamic}. 
\textcolor{black}{The computed flash-only image without alignment contains obvious artifacts, which can hardly provide effective cues for reflection removal. As a comparison, the flow estimation network can find the correspondences for objects with moderate motion, and then the two input images can be aligned to get a reflection-free flash-only image. As a result, the reflection can be well removed from the input ambient image with the guidance of the aligned flash-only image. } 
However, in extreme cases where the transmission changes rapidly in a very short time, we believe that it would not follow our assumption, and our approach may not be effective.} 

\textbf{{Remaining Limitations}
}
The reflection-free cue is based on the quality of the flash-only image. If all objects in transmission are too far and are not illuminated by the flash, there would be no difference between the ambient image and the flash image (i.e., the flash-only image will be totally black except for reflected flash) due to the irradiance falloff problem in Eq.~\ref{eq:Falloff}. In this case, our model will be degraded to single-image reflection removal. We believe other methods can be proposed to solve these cases.

\subsection{Study on Linear RGB Images}
\label{subsec:linear}

\textcolor{black}{In our experiments section, we show that the utilization of reflection-free flash-only images is vital in achieving the advancements in our method. In this part, we are interested in studying a problem: why such a human-designed cue can yield such noticeable enhancements, whereas neural networks cannot autonomously learn it from the data. In Section~\ref{sec:Architecture}, we argue that it is because the linearity does not exist in sRGB images after non-linear ISP operation. Using a pair of flash/no-flash raw images to compute a flash-only cue provides more information (i.e., the raw data and linearity information) than using a pair of flash/no-flash sRGB images. Hence, using the same training setting, replacing $I_{fo}$ with $I_{f}$ cannot achieve similar satisfying performance, as shown in Table~\ref{table:Ablation study}.
}

\textcolor{black}{The question that follows is: If we use images where linearity holds, is it necessary to manually compute a flash-only image? Can similar results be achieved from the flash/no-flash pair without manually computing a flash-only image? To answer the question, we design a framework that directly takes a pair of flash/no-flash images on linear space as input, as shown in Figure~\ref{fig:ArchitectureLinear}. For the new framework, we want to compare it with a baseline: we conduct the subtraction operation to obtain a flash-only linear RGB image to replace the flash image.}

\textcolor{black}{According to our experiments, we notice that computing a flash-only image does not improve the performance \textbf{on the linear image space}. Note that this result is consistent with our previous observation: it demonstrates the importance of using the linearity property of raw images. If the input images do not have linearity (e.g., sRGB images), we need to compute the flash-only image on raw space and then convert it to an sRGB image to provide guidance. If the input images have linearity (e.g., images in linear space), we do not need to compute the flash-only image. 
}

\section{Flash-only Reflection Removal Dataset}
\label{sec:Dataset}
\begin{figure}[t]
	\centering
	\begin{tabular}{@{}c@{\hspace{2mm}}c@{}}
		\includegraphics[width=0.48\linewidth]{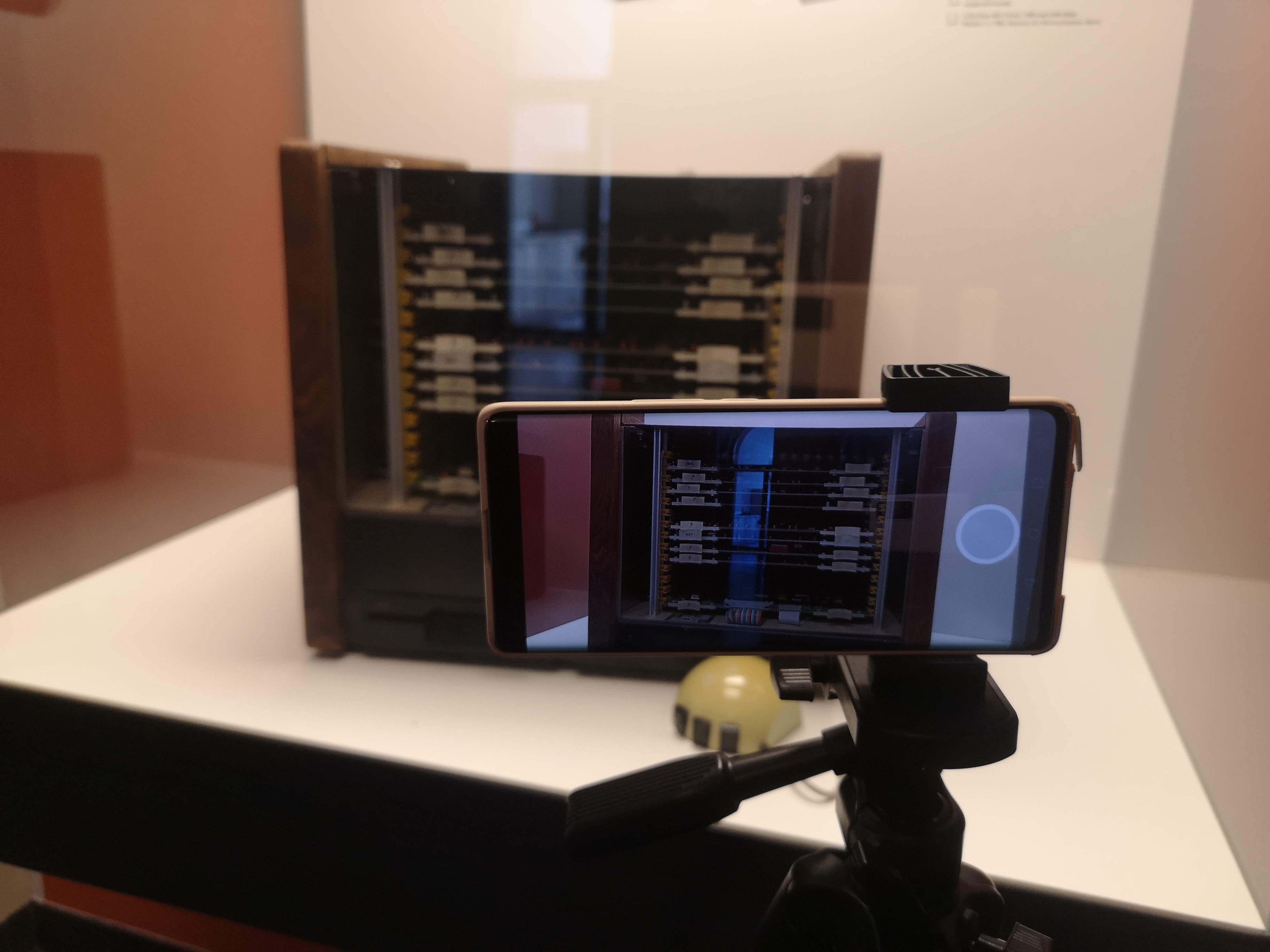} &

		\includegraphics[width=0.48\linewidth]{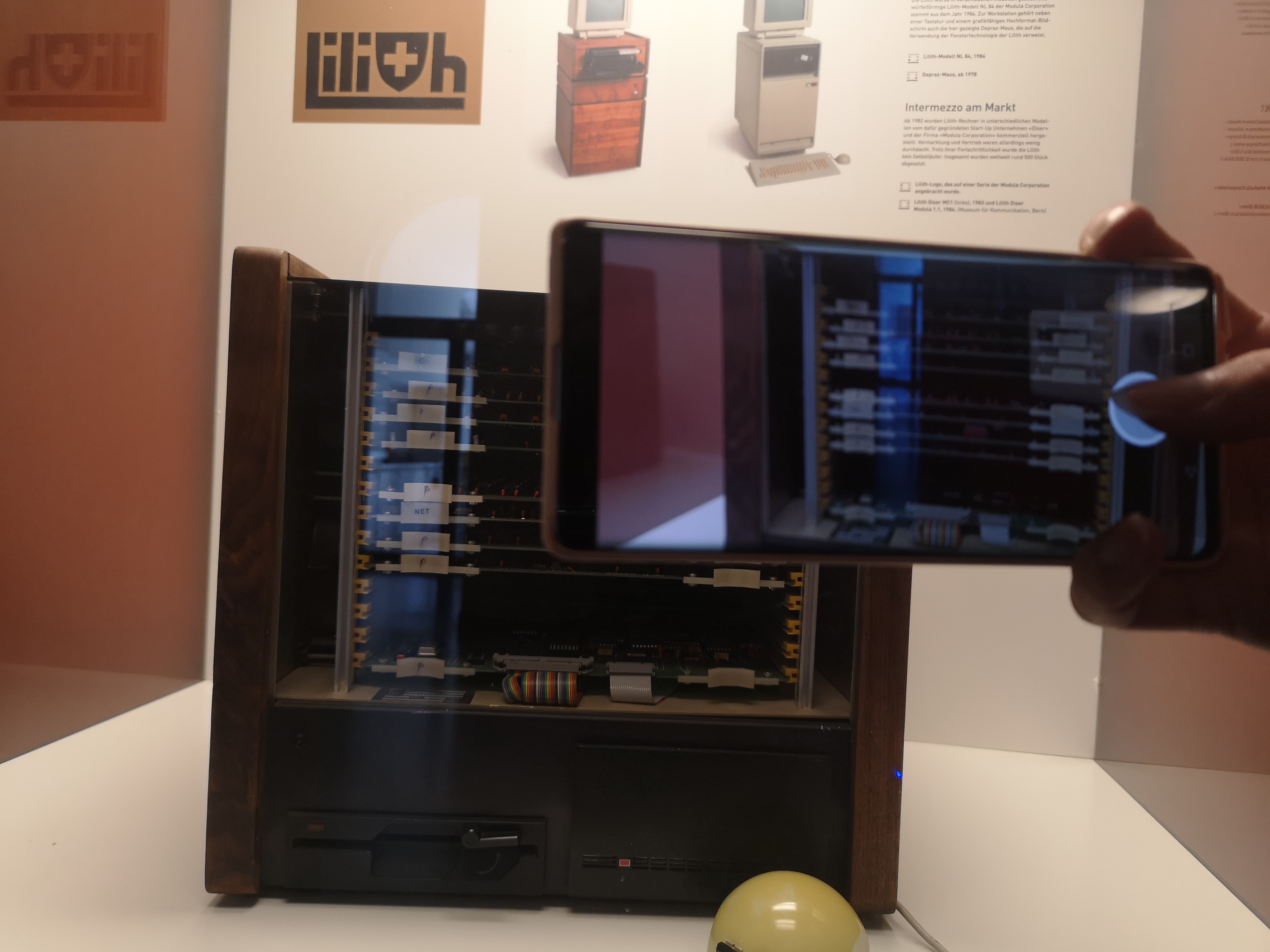} \\
		(a) Previous aligned pair~\cite{Lei_2021_RFC}       & (b) Current misaligned pair \\
	\end{tabular}
 \begin{tabular}{c}
       \includegraphics[width=0.94\linewidth]{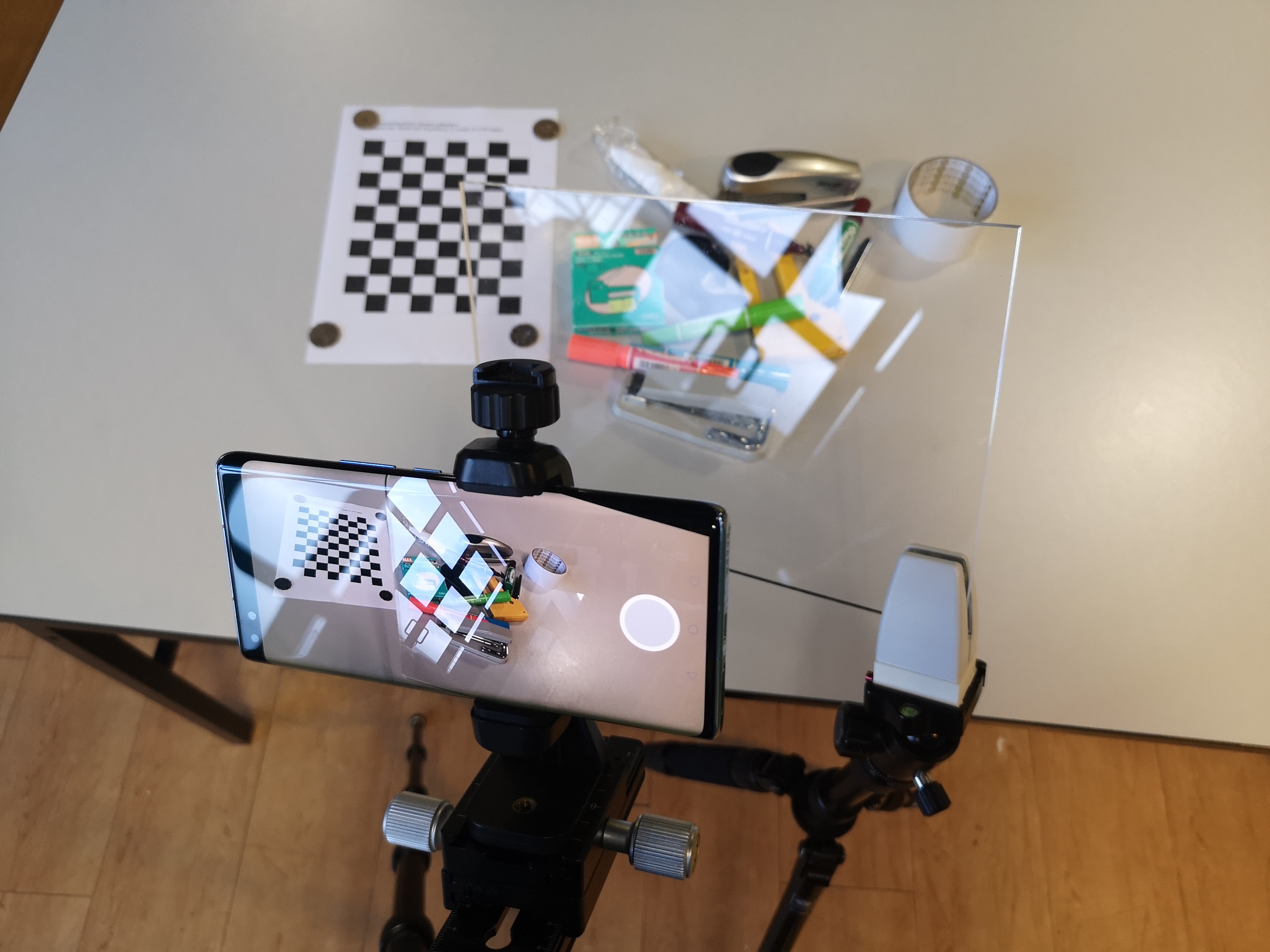} \\
       (c) Misaligned pair with ground truth
 \end{tabular}

	\vspace{1mm}
	\caption{Our data acquisition setup. A user only needs to click a button to capture a pair of flash/ambient images. In the previous test setting (a), we use a tripod to capture the aligned input pair images. However, in the current test setting (b), we directly hold the camera by hand to capture misaligned pair images. To evaluate our method quantitatively, we use a tripod in setting (c) to ensure perfect alignment between the ground truth and ambient images. We move the camera between the ambient shot and flash shot to deliberately introduce misalignment between the input images. 
}
	\label{fig:datacapcompare}
\end{figure}

\textcolor{black}{In this part, we introduce our FOR (Flash-Only Reflection-removal) datasets. The FOR dataset is composed of two subsets: FOR-Aligned and FOR-Misaligned datasets. The FOR-Aligned dataset targets at cases where the camera is fixed by a tripod and the flash and no-flash pair is perfectly aligned, while the FOR-Misaligned dataset focuses on the scenarios of handheld photography where the flash and no-flash pair is misaligned. 
}
\subsection{\textcolor{black}{FOR-Aligned Dataset}}
\subsubsection{Real-world data}
Our method requires a pair of raw flash/ambient images. Since there is no existing dataset, we construct the first real-world dataset that contains raw data for flash-based reflection removal. This dataset is collected by Nikon Z6 and a smartphone camera Huawei Mate30. We control the camera setting (e.g., exposure) to make sure that Eq.~\ref{eq:pureflash} holds. The collection procedure is as follows:

1) Fix the focal length, aperture, exposure time, and ISO.

2) Take the ambient image $I_a$ ($I_a^{raw}$).

3) Turn on the flash and take the flash image $I_f$ ($I_f^{raw}$).

4) To get the ground truth $T_a$, we turn off the flash and take an extra reflection image $R_a$ ($R_a^{raw}$). Note that this step is unnecessary at test time.

To collect high-quality data with perfect alignment, we use a tripod to fix the camera, as shown in Figure~\ref{fig:datacapcompare}. In practice, steps (1)-(3) can be programmed to be implemented automatically with a single shutter press on mobile phones. By doing so, the extra cost is mainly an extra shot. \textcolor{black}{Note that since we do not need to move the glass in our capture setting, we can utilize the diverse glass in the wild to construct our dataset.
}

We collect ground truth ambient transmission $T_a$ for training and evaluation. Specifically, we obtain $T_a^{raw} = I_a^{raw} - R_a^{raw}$ in raw (linear) space~\cite{Lei_2020_CVPR}. Thus, an extra reflection image under ambient illumination is captured. Then, ISP is implemented for each raw image, similar to the processing pipeline in Section~\ref{sec:data_prepare}. We adopt the metadata of $I_a^{raw}$ to process the raw data $T_a^{raw}$. At last, we crop the area where the transmission is valid following Lei et al.~\cite{Lei_2020_CVPR}. Briefly speaking, we capture a set $\{I_a^{raw}, I_f^{raw}, R_a^{raw}\}$
and process these three images to get the set $\{I_a, I_f, I_{fo}, T_a, R_a\}$. In total, we collect 157 sets of real-world images.

\subsubsection{Synthetic data} 
\label{subsubsec:aligned_synthetic}
Since the real-world dataset cannot provide enough data for training, we construct an extra synthetic dataset by the composition of transmission and reflection layers in raw or linear space $I_a^{raw}=R_a^{raw}+T_a^{raw}$. The transmission layers of flash-only and ambient illumination are obtained from a flash dataset~\cite{aksoy2018ECCV_flashdataset} with 1964 raw ambient images $T_a$ and flash-only images $I_{fo}$ pairs. Two kinds of reflections $R_a$ are provided for each $T_a$ to synthesize the ambient image $I_a$. The first type of reflection is another ambient image in the flash dataset that is quite sharp. Then real-world reflections collected by Wan et al.~\cite{wan2019corrn} are used as the second type of reflection, since there are many blurry reflections and few sharp reflections in their dataset~\cite{wan2019corrn}. We reverse gamma correction to mimic the raw data because raw data is not provided in this dataset. The transmission and reflectance are dimmed according to transmission and reflectance in raw space, which we fixed at 0.61 and 0.22.
Most images in the flash dataset are $1960 \times 1080$, while all the images in the real-world reflection dataset are $353 \times 506$. Therefore, we resize images in the flash dataset to $\frac{1}{2}$ with bilinear interpolation in linear space. We also apply data augmentation by random cropping transmission and reflection before composition with a scale of 0.8. In addition, the final crop sizes are rounded to multiples of 32 for the convenience of downsampling operations in UNet and PWC-Net.

\subsubsection{Dataset split}
For the real-world dataset, we use 77, 30, and 50 sets of images for training, validation, and evaluation. There is no overlapping reflection or transmission between the training and test sets. The synthetic data is only used as a supplement for training since the real-world reflection images in CoRRN~\cite{wan2019corrn} are in a chaotic order, and no dataset split is available.

\begin{table*}[t]
	\small
	\caption{Quantitative comparison results among our method and previous methods on \textcolor{black}{FOR-Aligned dataset}. }
	\label{table:Metrics}
	\centering
	\renewcommand{\arraystretch}{1.2}

	\begin{tabular}{cc@{\hspace{5.1mm}}c@{\hspace{5.1mm}}c@{\hspace{5.1mm}}c@{\hspace{5.1mm}}c@{\hspace{5.1mm}}c@{\hspace{5.1mm}}c@{\hspace{5.1mm}}c@{\hspace{5.1mm}}c@{\hspace{5.1mm}}}

		\toprule[1pt]

		                  & Input $I_a$       & {Zhang}                       & BDN                      & Wei et al.                & Kim et al.            & Li et al.            & {Agrawal}                               & {SDN}                    & Ours           \\

		                  &                   & et al.~\cite{zhang2018single} & ~\cite{eccv18refrmv_BDN} & ~\cite{wei2019single_ERR} & ~\cite{Kim_2020_CVPR} & ~\cite{Li_2020_CVPR} & et al.~\cite{agrawal2005removing_flash} & ~\cite{chang2020siamese} &                \\

		\midrule

		\#Input images    & 1                 & 1                             & 1                        & 1                         & 1                     & 1                    & 2                                       & 2                        & 2              \\
		\midrule
		PSNR$\uparrow$    & 22.72             & 23.76                         & 21.41                    & 23.89                     & 21.67                 & \underline{24.53}    & 23.13                                   & 22.63                    & \textbf{29.76} \\
		SSIM$\uparrow$    & 0.874             & 0.873                         & 0.802                    & 0.864                     & 0.821                 & \underline{0.890}    & 0.853                                   & 0.827                    & \textbf{0.930} \\
		LPIPS$\downarrow$ & \underline{0.205} & 0.242                         & 0.410                    & 0.238                     & 0.298                 & {0.224}              & 0.251                                   & 0.269                    & \textbf{0.156} \\

		\bottomrule[1pt]
	\end{tabular}
	\vspace{1mm}
\end{table*}

\begin{figure*}[t!]
	\centering
	\begin{tabular}{@{}c@{\hspace{1mm}}c@{\hspace{1mm}}c@{\hspace{1mm}}c@{\hspace{1mm}}c@{\hspace{1mm}}c@{}}

		\includegraphics[width=0.16\linewidth]{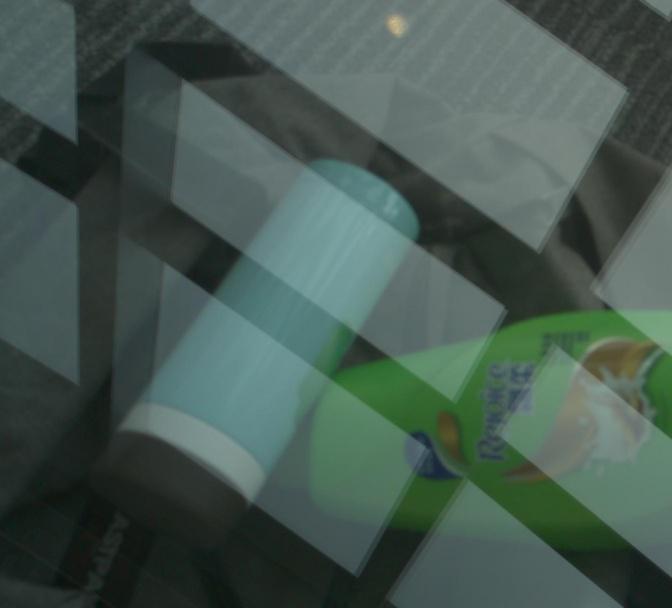}   &
		\includegraphics[width=0.16\linewidth]{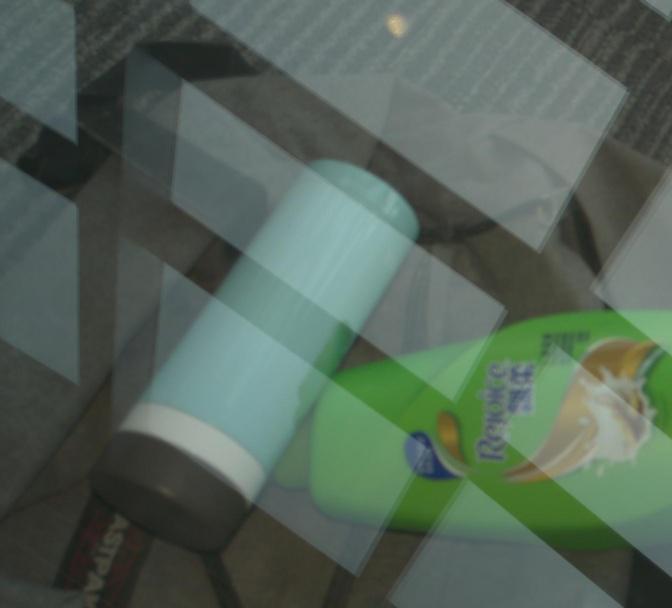}     &
		\includegraphics[width=0.16\linewidth]{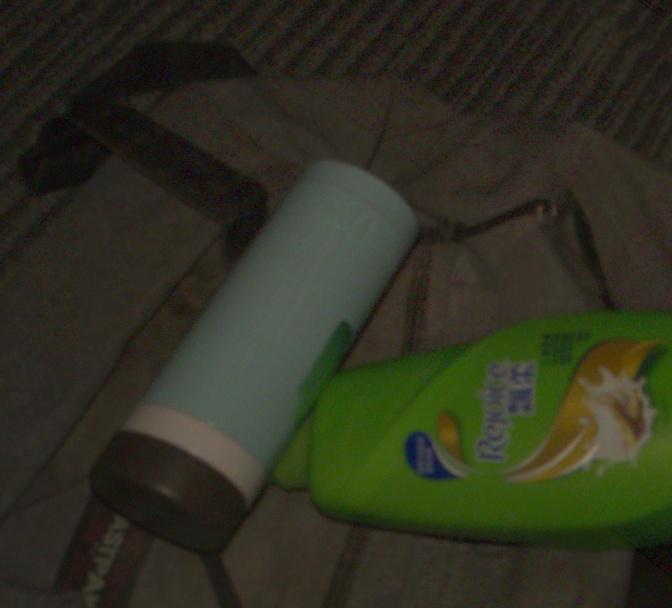} &
		\includegraphics[width=0.16\linewidth]{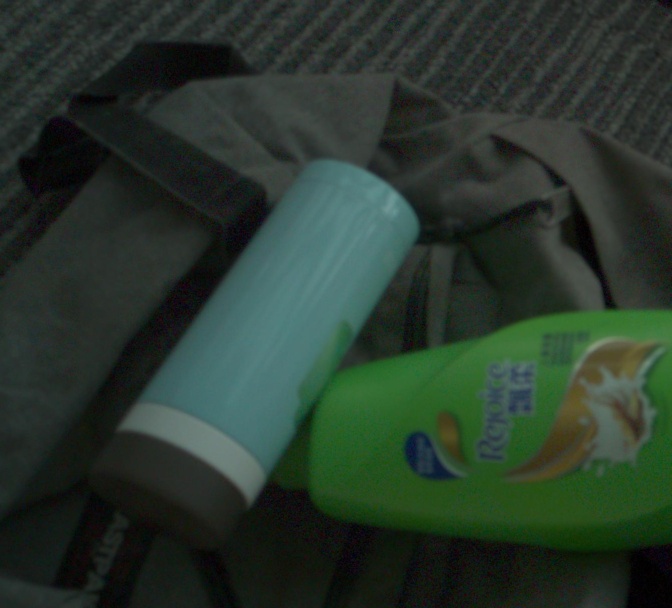}   &
  		\includegraphics[width=0.16\linewidth]{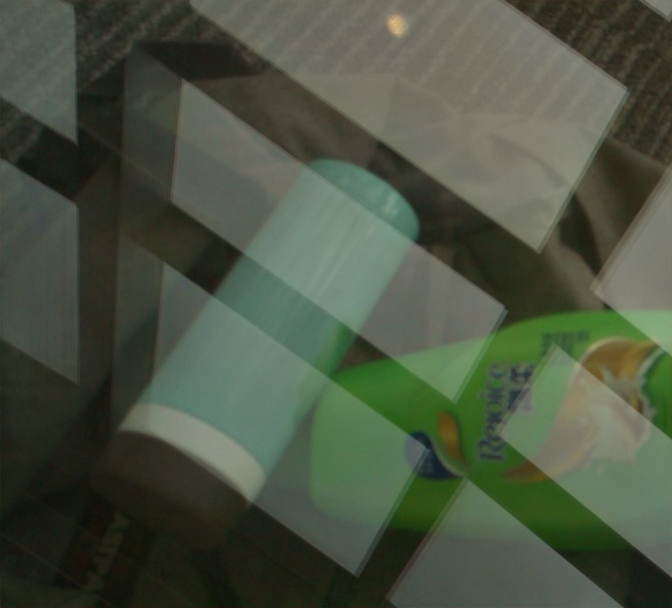}                          &
		\includegraphics[width=0.16\linewidth]{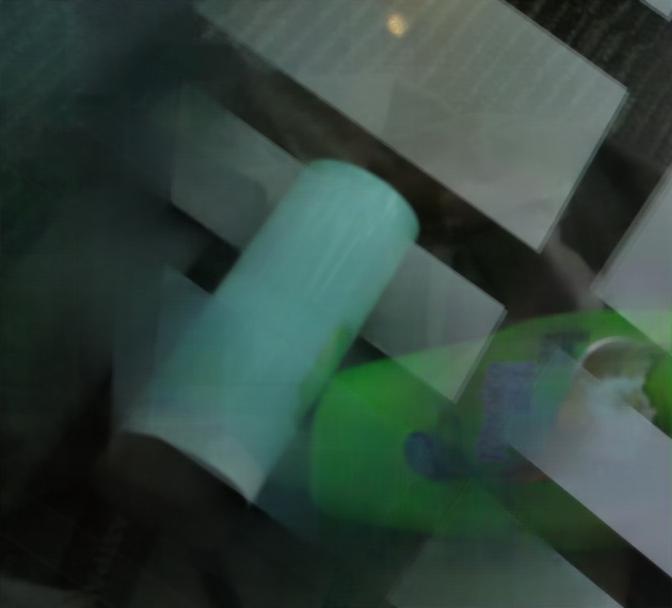}                            \\               
				Input $I_a$ & Input $I_f$                     & Processed $I_{fo}$            & Ground truth  &    Zhang et al.~\cite{zhang2018single} & BDN~\cite{eccv18refrmv_BDN} 	\\    
		\includegraphics[width=0.16\linewidth]{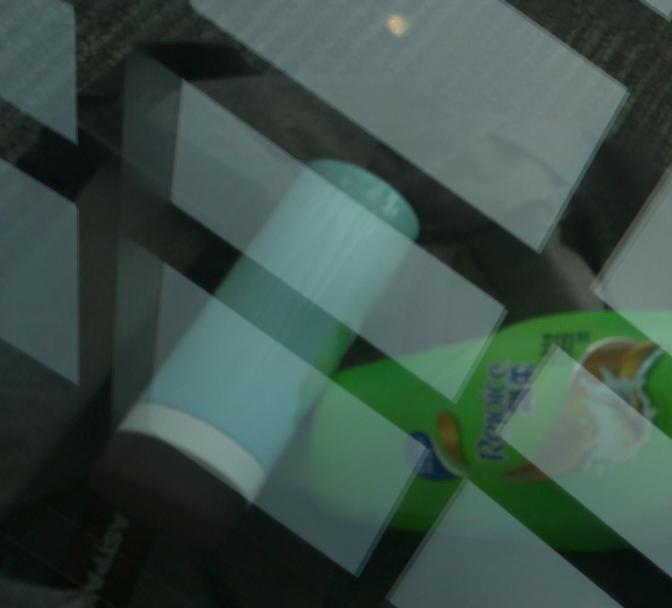}                           &
		\includegraphics[width=0.16\linewidth]{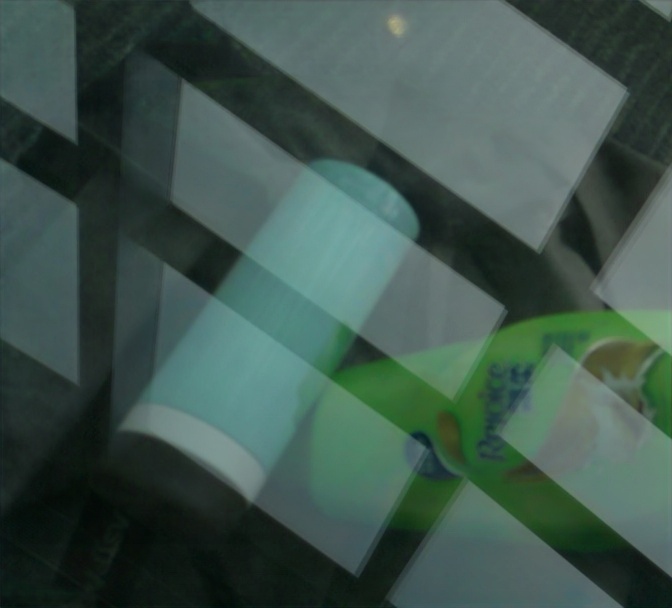}                            &
		\includegraphics[width=0.16\linewidth]{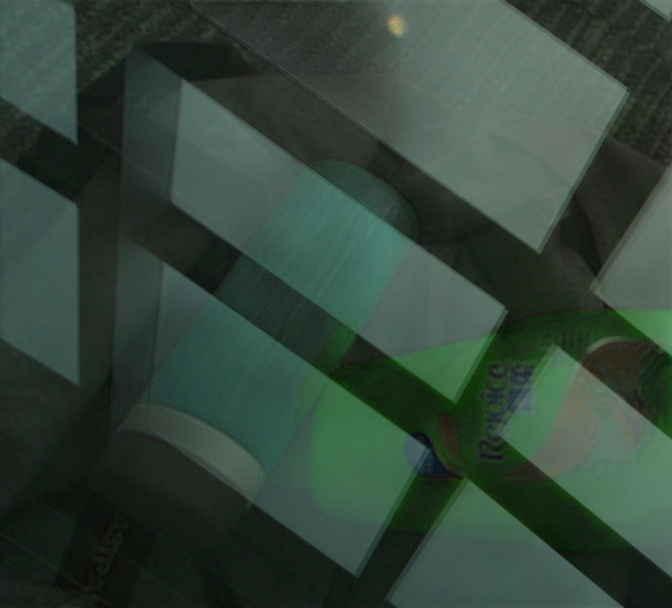}                          &
		\includegraphics[width=0.16\linewidth]{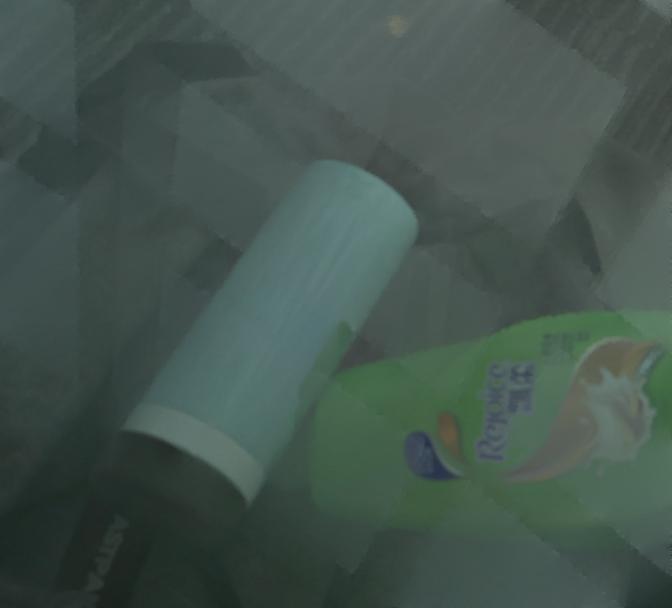}                             &
		\includegraphics[width=0.16\linewidth]{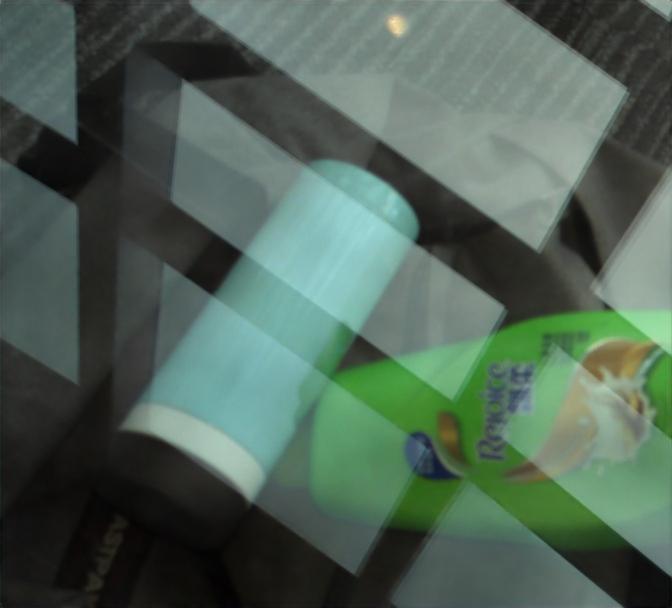}                            &
		\includegraphics[width=0.16\linewidth]{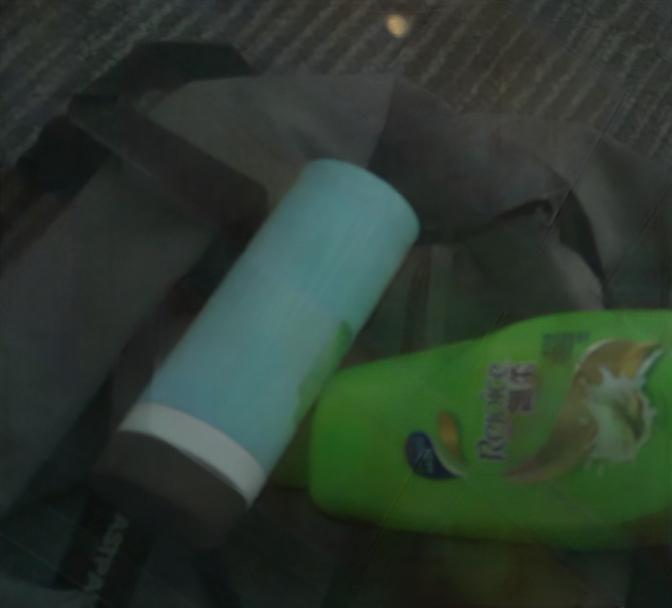}                                                                                                                                                                                         \\

                               	Wei et al.~\cite{wei2019single_ERR}                                                                            & Kim et al.~\cite{Kim_2020_CVPR} & 
Li et al.~\cite{Li_2020_CVPR} & {Agrawal et al.~\cite{agrawal2005removing_flash}} & SDN~\cite{chang2020siamese}         & Ours                        \\
	\end{tabular}
	\vspace{1mm}
	\caption{Qualitative comparison to baselines on a real-world image that contains strong reflection.}
	\label{fig:StrongComparison}
\end{figure*}

\subsection{\textcolor{black}{FOR-Misaligned Dataset}}
\label{subsec:misalign_data}

\subsubsection{Real-world data}
For real-world handheld photography, we develop an Android App to capture a burst of raw no-flash and flash input pair and we collect the dataset using the smartphone camera on Honor V40. However, aligned ground truth is difficult to obtain without using a tripod. Therefore we collect a misalignment dataset with aligned ground truth according to statistics of handheld motion. We first estimate handheld motion statistics during burst by capturing images of a static checkerboard of known size. According to our estimation, the translations are generally within 5mm, and the rotations are usually within 1 degree. Then we collect several $I_a$, $I_f$, $R_a$ groups of the same scene with small misalignment, fixed focal length and aperture, where a static checkerboard is put in the view to estimate the pose of each group. Then we sample group pairs $I_{1a}$, $I_{1f}$, $R_{1a}$; $I_{2a}$, $I_{2f}$, $R_{2a}$ according to their relative motion and $I_{1a}$,$R_{1a}$,$I_{2f}$ are selected as data. Before the preprocessing steps, we crop $I_{1a}$,$R_{1a}$,$I_{2f}$ with the same bounding box, where the checkerboard is not visible in each image. Finally, 114 sets of images are used for evaluation.

\subsubsection{Synthetic data}
The collection of real misalignment data is time-consuming. Hence, we also construct a synthetic dataset for training following the pipeline mentioned in Section~\ref{subsec:misalign}. As for the source data for synthesizing the images, we follow the same setting of our aligned dataset in Section~\ref{subsubsec:aligned_synthetic}. Besides, we also try another type of synthetic misalignment \textit{for comparison}. We generate random homography transformations by randomly perturbing the 4 endpoints of the image with maximum displacement 8 and estimating the homography for the 4 endpoints. For each group of data, we generate 2 different homography $H_T$ and $H_R$ for transmission and reflection layers respectively. Then we warp $T_a$, $I_{fo}$ with $H_T$, warp $R_a$ with $H_R$ and merge the warped images to obtain the new flash image $I_f$.

\begin{figure*}[]
	\centering
	\begin{tabular}{@{}c@{\hspace{1mm}}c@{\hspace{1mm}}c@{\hspace{1mm}}c@{\hspace{1mm}}c@{}}

		\includegraphics[width=0.193\linewidth]{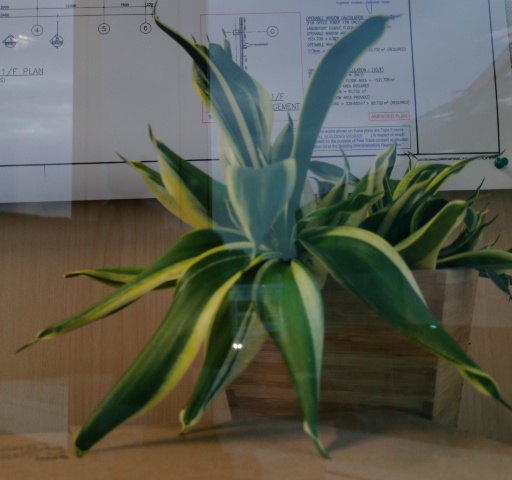}                            &
		\includegraphics[width=0.193\linewidth]{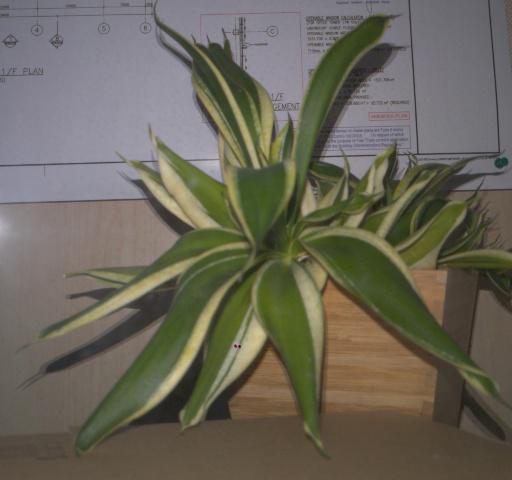}   &
		\includegraphics[width=0.193\linewidth]{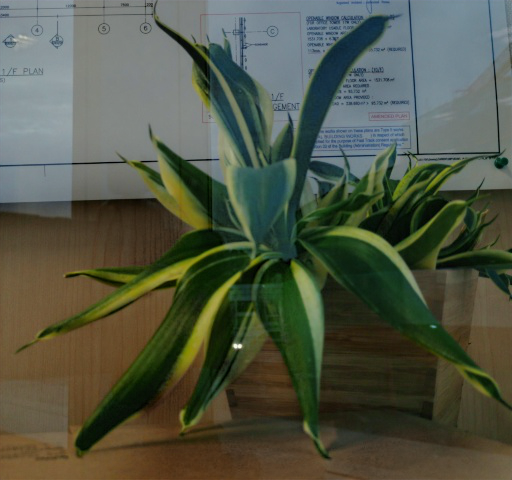}   &
		\includegraphics[width=0.193\linewidth]{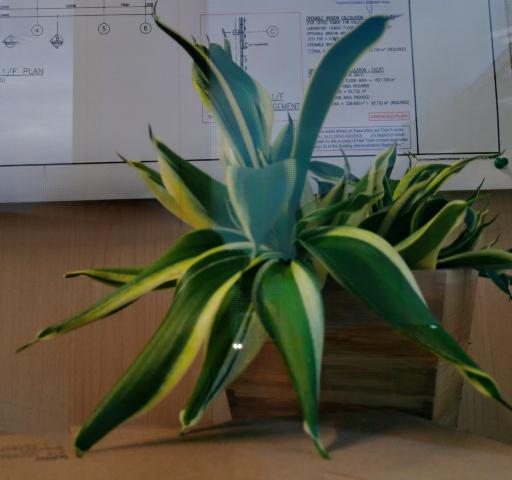}              &
		\includegraphics[width=0.193\linewidth]{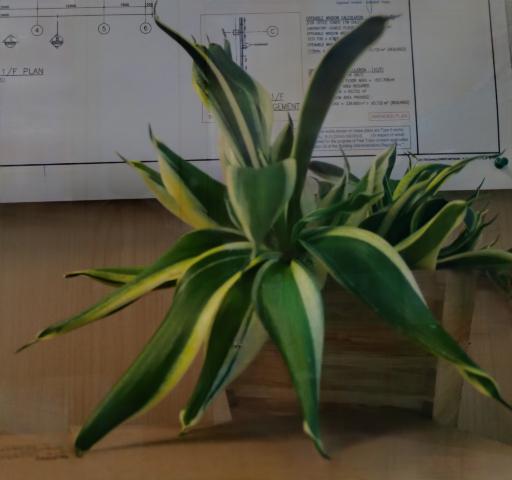}                                                                                                                               \\

		\includegraphics[width=0.193\linewidth]{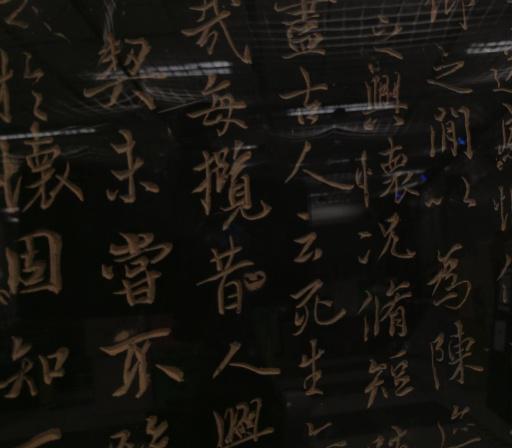}   &
		\includegraphics[width=0.193\linewidth]{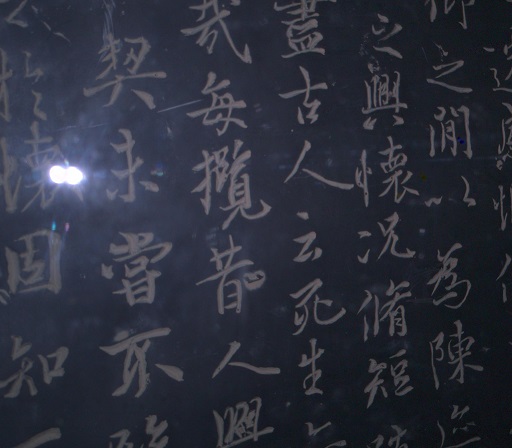}          &
		\includegraphics[width=0.193\linewidth]{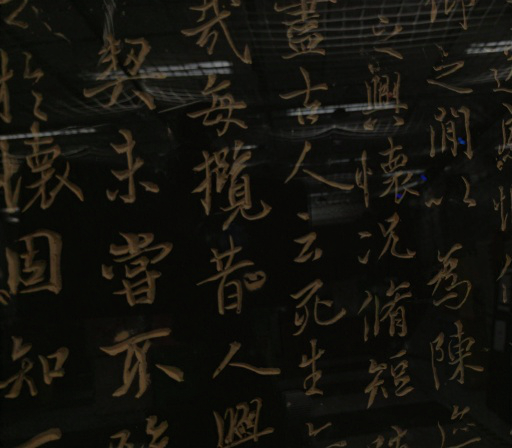} &
		\includegraphics[width=0.193\linewidth]{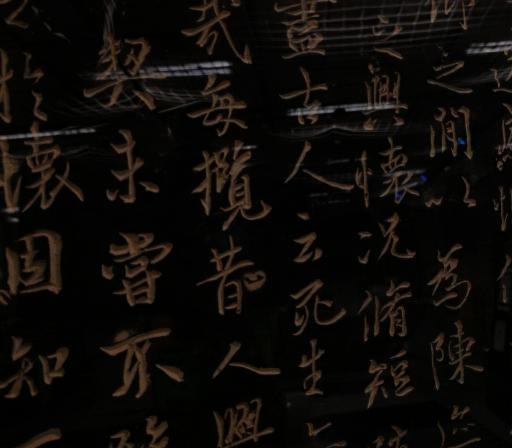}            &
		\includegraphics[width=0.193\linewidth]{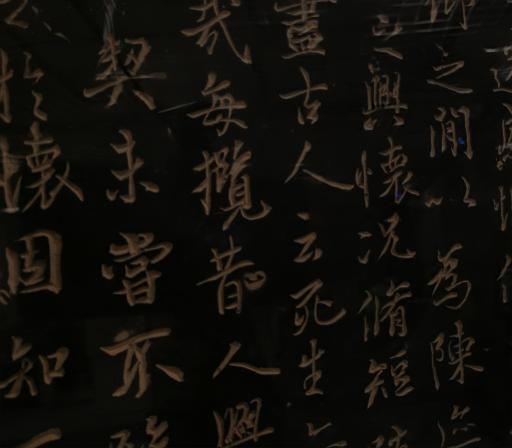}                                                                                                 \\

		Input ambient image                                                                                             & Processed $I_{fo}$ & Li et al.~\cite{Li_2020_CVPR} & Wei et al.~\cite{wei2019single_ERR} & Ours \\
	\end{tabular}
	\vspace{1mm}
	\caption{Qualitative comparison to single-image based baselines~\cite{Li_2020_CVPR,wei2019single_ERR} on real-world images.}
	\label{fig:ComparisonSingle}
\end{figure*}

\begin{figure}[t]
	\centering
\begin{tabular}{@{}c@{\hspace{1mm}}c@{\hspace{1mm}}c@{\hspace{1mm}}c@{}}
&
\includegraphics[width=0.321\linewidth]{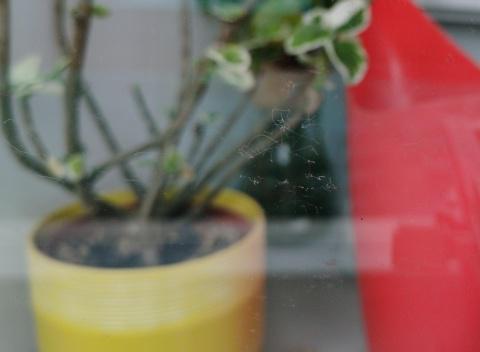}&
\includegraphics[width=0.321\linewidth]{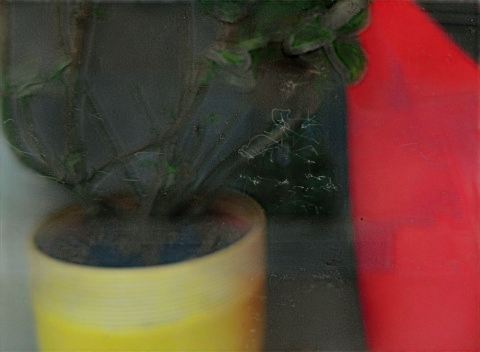}&
\includegraphics[width=0.321\linewidth]{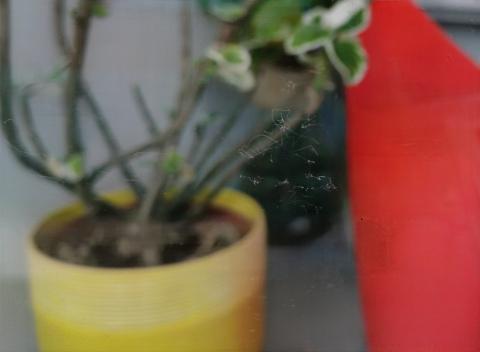}\\
&Input $I_{a}$ & Li et al.~\cite{Li_2020_CVPR} &  Ours
\end{tabular}
	\vspace{1mm}
	\caption{Comparison to Li et al.~\cite{Li_2020_CVPR} on a real-world image that contains blurry transmission. }
	\label{fig:ComparisonBlur}
\end{figure}

\section{Experiments}
\subsection{Experimental Setup}
We introduce the datasets and baselines in this part.

\textcolor{black}{
\textbf{Dataset.} We mainly use two real-world datasets in this section: FOR-Aligned dataset and FOR-Misaligned dataset. The previous version of this paper mainly uses the FOR-Aligned dataset, which requires the input pair of ambient and flash images are perfectly aligned. In this version, most experiments are conducted on FOR-Misaligned dataset. We refer the readers to Section~\ref{sec:Dataset} for the details of dataset.  }

\textbf{Baselines.} We first select two flash-based reflection removal methods: Agrawal et al.~\cite{agrawal2005removing_flash} and SDN~\cite{chang2020siamese}. Then we select several single-image methods for comparison, including Zhang et al.~\cite{zhang2018single}, Wei et al.~\cite{wei2019single_ERR}, BDN~\cite{eccv18refrmv_BDN}, Li et al.~\cite{Li_2020_CVPR}, and Kim et al.~\cite{Kim_2020_CVPR}. For Agrawal et al.~\cite{agrawal2005removing_flash}, we observe that it is wrongly used in the comparison of SDN~\cite{chang2020siamese}: they use the flash image instead of the flash-only image as guidance; in our comparison, we adopt the flash-only image as the input to Agrawal et al.~\cite{agrawal2005removing_flash}. For SDN~\cite{chang2020siamese}, we use predicted ambient transmission for quantitative comparison. We retrain the models whose training codes are available on our constructed training set and choose the better results between pretrained models and retrained models.

\subsection{Comparison to Baselines}

In Table~\ref{table:Metrics}, we adopt PSNR, SSIM, and LPIPS~\cite{zhang2018lpips} as quantitative evaluation metrics, and our model obtains the best scores on all metrics. Specifically, our method outperforms state-of-the-art reflection removal approaches by more than 5.23dB in PSNR, 0.04 in SSIM, and 0.068 in LPIPS on the real-world dataset.

In Figure~\ref{fig:StrongComparison}, we compare our approach with all mentioned baselines. Both single-image baselines~\cite{Kim_2020_CVPR,Li_2020_CVPR,wei2019single_ERR,eccv18refrmv_BDN, zhang2018single} and flash-based baselines~\cite{agrawal2005removing_flash,chang2020siamese} cannot correctly remove reflection. As can be seen, our approach can remove very strong reflection and recover underlying transmission. It is because processed flash-only image $I_{fo}$ is still reflection-free for strong reflection, and thus provides strong guidance.

In Figure~\ref{fig:ComparisonSingle}, we further compare our method with single-image methods~\cite{Li_2020_CVPR,wei2019single_ERR} that obtain quantitative scores.
In the first row, the edge of the reflection is sharp. The second row is a picture of calligraphy writing, in which both reflection and transmission have rare semantic information. As can be seen, two single-image methods~\cite{Li_2020_CVPR,wei2019single_ERR} cannot remove the reflections.
Our method removes reflections well since the reflection-cue is independent of the appearance (e.g., smoothness and semantic information) of reflection.

In Figure~\ref{fig:ComparisonBlur}, we compare with Li et al.~\cite{Li_2020_CVPR} on an image that contains blurry transmission. The result of  Li et al.~\cite{Li_2020_CVPR} removes transmission wrongly since their method cannot distinguish the reflection correctly. As a comparison, our approach can easily distinguish the reflection and avoid removing transmission wrongly because the reflection-free cue is independent of smoothness.

\begin{figure*}[h]
	\centering
	\begin{tabular}{@{}c@{\hspace{1mm}}c@{\hspace{1mm}}c@{\hspace{1mm}}c@{\hspace{1mm}}c@{\hspace{1mm}}c@{}}
		                                                                                        &
		\includegraphics[width=0.191\linewidth]{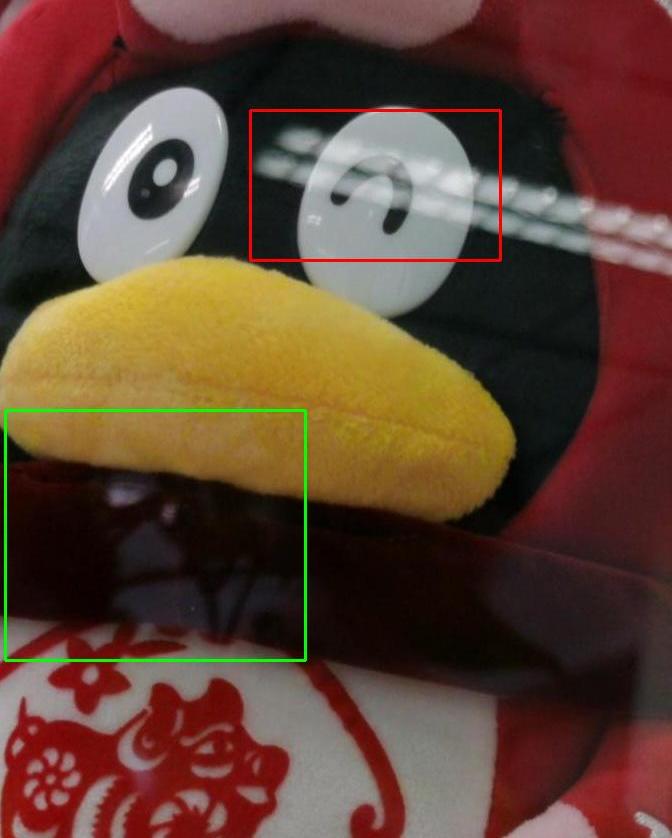} &
		\includegraphics[width=0.191\linewidth]{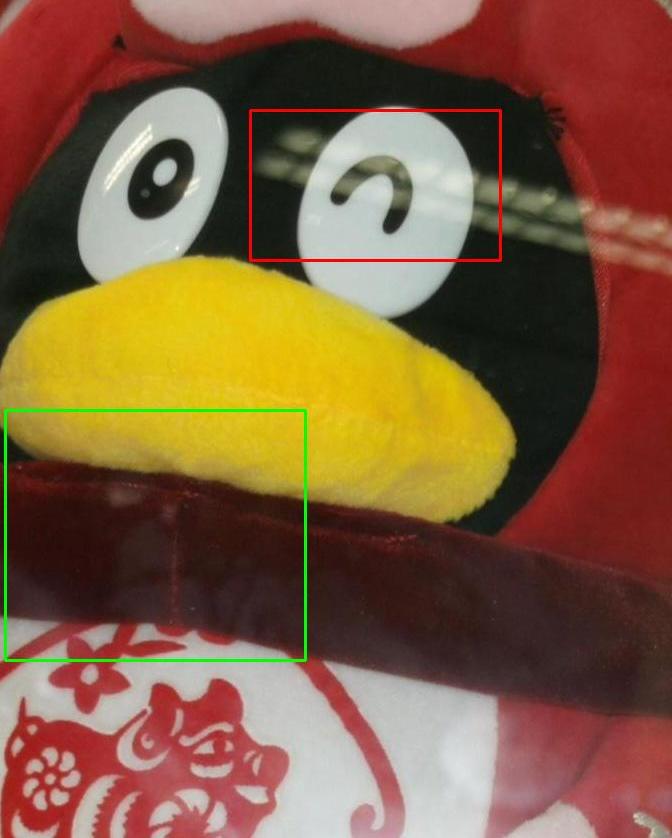} &
		\includegraphics[width=0.191\linewidth]{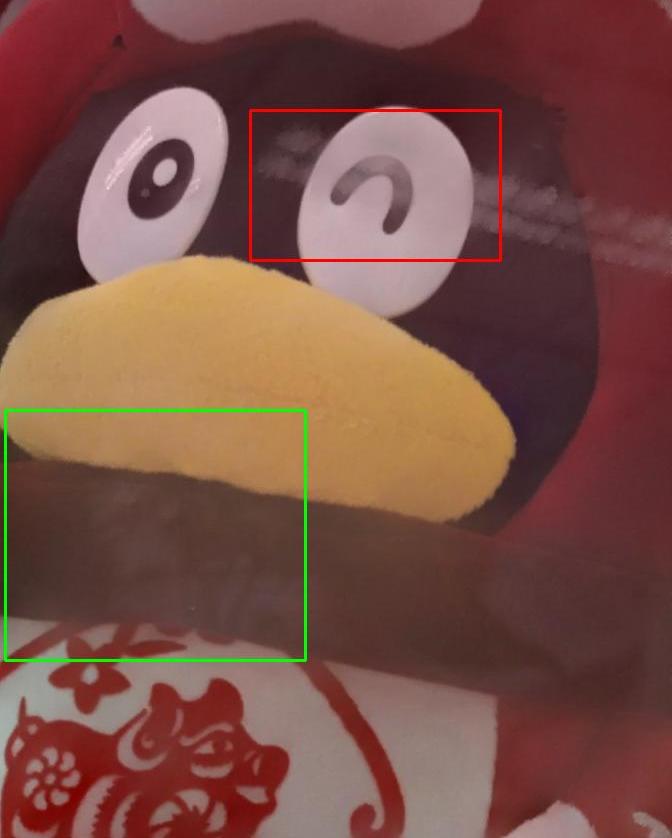}    &
		\includegraphics[width=0.191\linewidth]{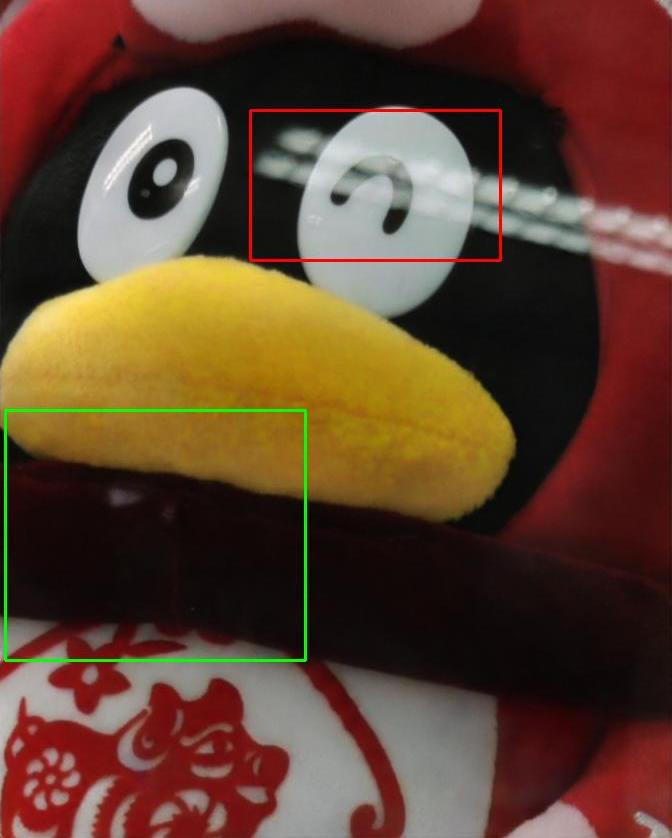} &
		\includegraphics[width=0.191\linewidth]{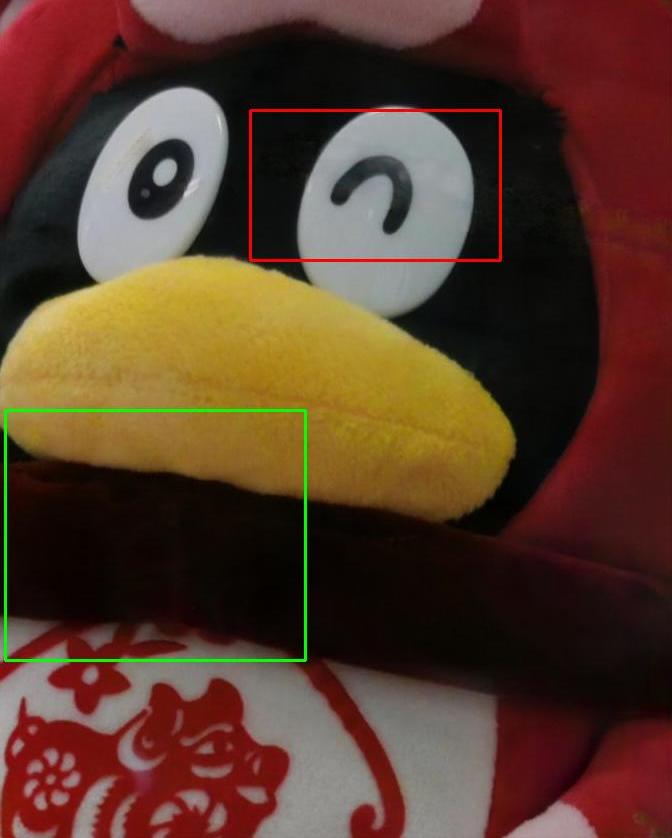}
		\\
		\rotatebox{90}{\small Weak reflection area}                                             &
		\includegraphics[width=0.191\linewidth]{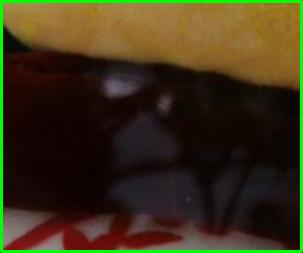}      &
		\includegraphics[width=0.191\linewidth]{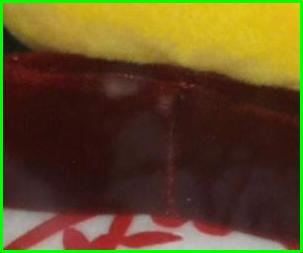}      &
		\includegraphics[width=0.191\linewidth]{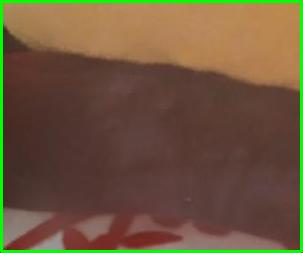}         &
		\includegraphics[width=0.191\linewidth]{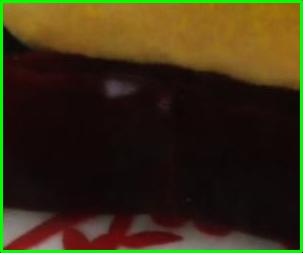}      &
		\includegraphics[width=0.191\linewidth]{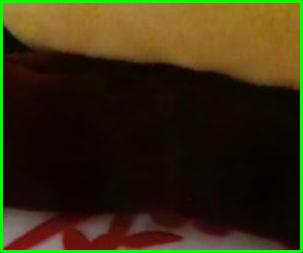}
		\\

& Input ambient image & Input flash image & Agrawal et al.~\cite{agrawal2005removing_flash} & SDN~\cite{chang2020siamese} & Ours \\
	\end{tabular}
	\vspace{1mm}
	\caption{Qualitative comparison to flash-based reflection removal baselines on real-world images. Results of Agrawal et al.~\cite{agrawal2005removing_flash} contain reflection residuals and are over-smooth. For SDN~\cite{chang2020siamese}, they can remove the weak reflection but cannot remove the strong reflection.}
	\label{fig:ComparisonFlash}
\end{figure*}

In Figure~\ref{fig:StrongComparison} and Figure~\ref{fig:ComparisonFlash}, we compare our method with two flash-based methods~\cite{agrawal2005removing_flash, chang2020siamese}.
The results of Agrawal et al.~\cite{agrawal2005removing_flash} usually remove too many details and cannot completely remove reflection. For SDN~\cite{chang2020siamese}, they can remove weak reflection but cannot remove strong reflection. It is because they require the reflection to be well suppressed, but the strong reflection cannot be suppressed by flash. Our method removes both weak and strong reflection. Also, the details and color are consistent with ambient images in our results.

\begin{figure*}
	\centering
	\begin{tabular}{@{}c@{\hspace{1mm}}c@{\hspace{1mm}}c@{\hspace{1mm}}c@{\hspace{1mm}}c@{\hspace{1mm}}c@{}}
		                                                                                                            &
		\includegraphics[width=0.193\linewidth]{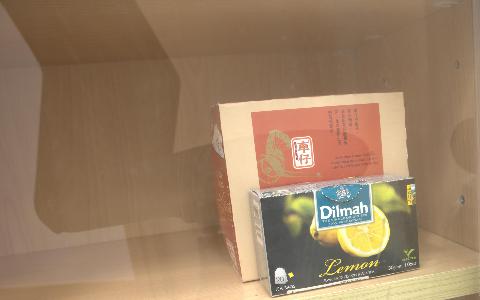}     &
		\includegraphics[width=0.193\linewidth]{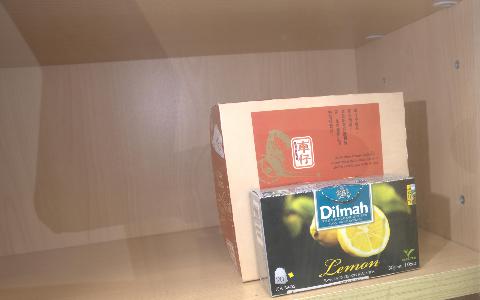}       &
		\includegraphics[width=0.193\linewidth]{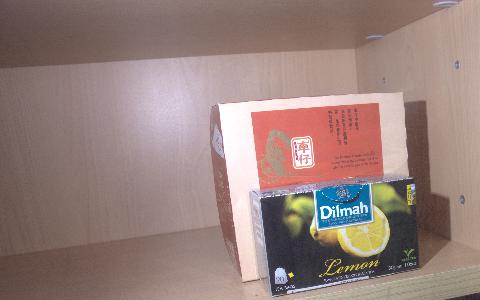}   &
		\includegraphics[width=0.193\linewidth]{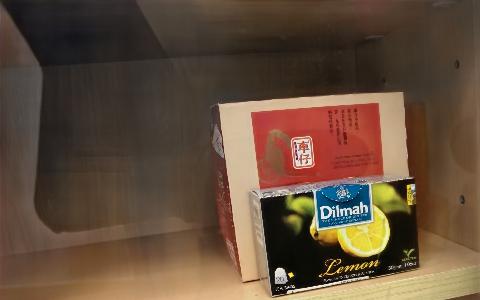} &
		\includegraphics[width=0.193\linewidth]{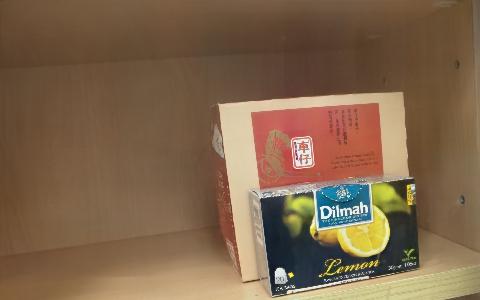}
		\\

		                                                                                                            &
		(a) Input $I_{a}$                                                                                           & (b) Input $I_{f}$ & (c) Processed $I_{fo}$ & (d) Our $\hat T$: using $I_f$ & (e) Our $\hat T$: using $I_{fo}$ \\
	\end{tabular}
	\vspace{1mm}
	\caption{Qualitative comparison between using $I_a+I_f$ and $I_a+I_{fo}$ as input. Note that the edge suppressed in flash image (b) is removed in (d). However, most reflections are not suppressed, and thus the perceptual quality of $(d)$ is not good.}
	\label{fig:AblFlash}
\end{figure*}

\begin{table}[]
	\small
  \caption{Quantitative comparison among our complete model and multiple ablated models of our methods on a \textcolor{black}{FOR-Aligned} dataset. }
  \label{table:Ablation study}
	\centering
	\renewcommand{\arraystretch}{1.2}
	\begin{tabular}{l@{\hspace{6mm}}lc@{\hspace{4mm}}c@{\hspace{4mm}}c@{\hspace{4mm}}}
		\toprule[1pt]

		Architecture & Input          & PSNR$\uparrow$    & SSIM$\uparrow$    & LPIPS$\downarrow$ \\
		\midrule
		$g_B$        & $I_a$+$I_{f}$  & 26.99             & 0.911             & 0.204             \\
		$g_B$        & $I_a$+$I_{fo}$ & \underline{27.55} & \underline{0.917} & \underline{0.187} \\
		$g_R + g_T$  & $I_a$          & 25.13             & 0.888             & 0.258             \\
		$g_R + g_T$  & $I_a$+$I_{f}$  & 27.21             & \underline{0.917} & 0.196             \\
		$g_R + g_T$  & $I_a$+$I_{fo}$ & \textbf{29.76}    & \textbf{0.930}    & \textbf{0.156}    \\
		\bottomrule[1pt]
	\end{tabular}
	\vspace{1mm}

\end{table}

\subsection{Ablation Study}
\label{sec:Analysis}

\subsubsection{Reflection-free cues}
Although it is quite simple to compute the reflection-free flash-only image, it can improve quantitative and qualitative results a lot. To demonstrate the importance of $I_{fo}$, we modify the input of the first network: (1) Replace $I_{fo}$ with $I_{f}$. (2) Use a single $I_a$ as input. Table~\ref{table:Ablation study} shows the quantitative results. Under the same training setting, using a single $I_a$ gets the worst scores, and replacing $I_{fo}$ with $I_f$ also degrades the performance.

We find that the weakness of using the flash image $I_f$ instead of the flash-only image $I_{fo}$ is similar to SDN~\cite{chang2020siamese} in the qualitative comparison. In Figure~\ref{fig:AblationStudy}, the reflection is not well suppressed by flash, but the flash-only image is still reflection-free. In this case, replacing $I_{fo}$ with $I_{f}$ performs poorly. Another example is also shown in Figure~\ref{fig:AblFlash}(d), replacing $I_{fo}$ by $I_f$ leads to obvious artifacts when reflection cannot be suppressed by the flash. Moreover, it cannot handle novel shadows brought by the flash.

\subsubsection{Reflection-pass network}
As introduced in Section~\ref{sec:Architecture}, the dedicated architecture is vital to avoid absorbing artifacts of flash-only images. In Figure~\ref{fig:AblationStudy}, the base model can remove the reflection well, but artifacts (e.g., color distortion) appear in the result. As a comparison, the result of the complete model does not contain obvious artifacts. In addition, using our dedicated architecture also improves the quantitative performance a lot, as shown in Table~\ref{table:Ablation study}.

Note that although `$g_B$, $I_a + I_{fo}$' has similar quantitative scores with `$g_R+g_T, I_{a}+I_{f}$', the reasons for degradation are different: we observe the former can remove most reflection but usually has artifacts of flash-only images; the latter generally cannot remove strong reflection correctly.

\begin{figure*}[h]
	\centering
	\begin{tabular}{@{}*{3}{c@{\hspace{1mm}}}c@{}}
		\includegraphics[width=0.243\linewidth]{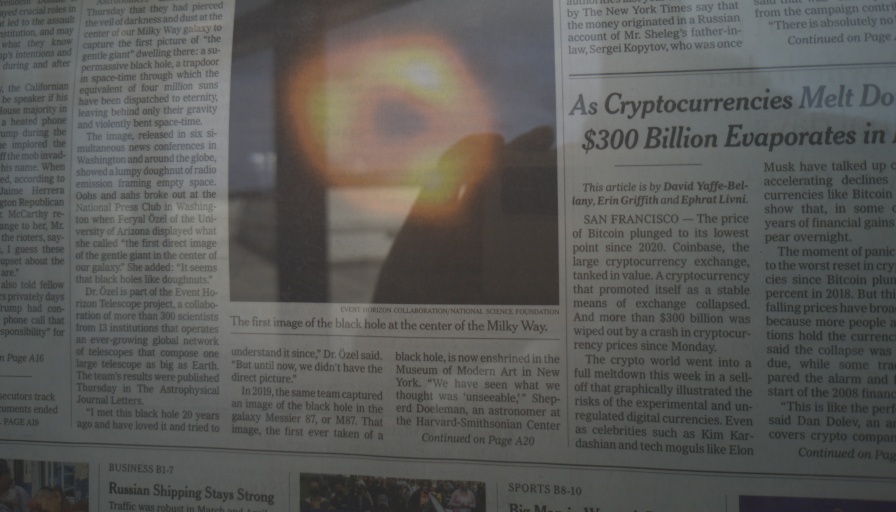}         &
		\includegraphics[width=0.243\linewidth]{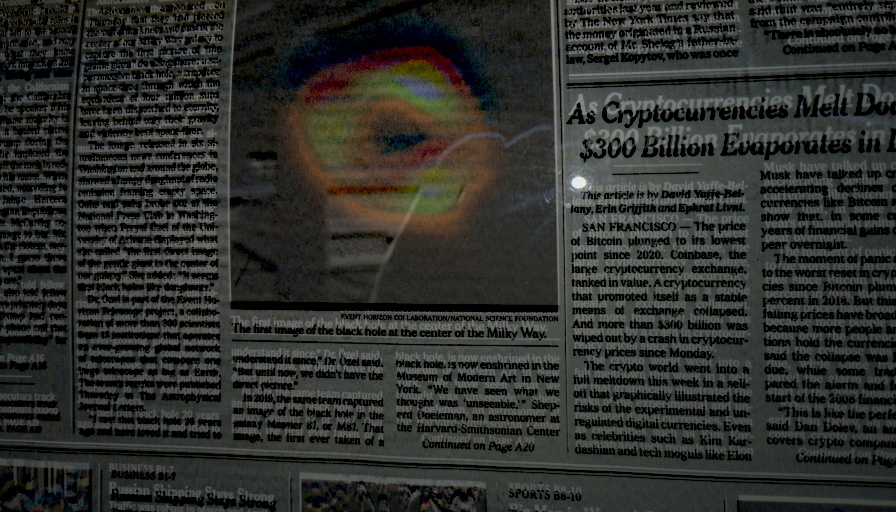}         &
		\includegraphics[width=0.243\linewidth]{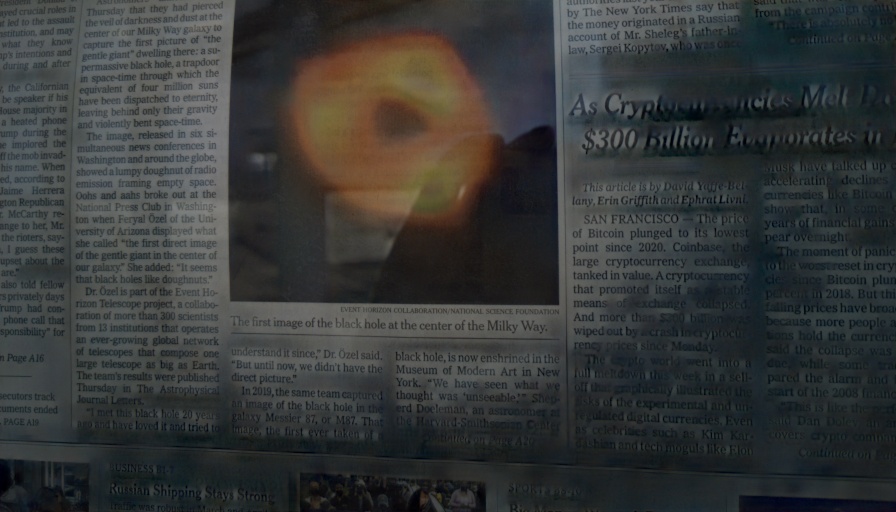} &
		\includegraphics[width=0.243\linewidth]{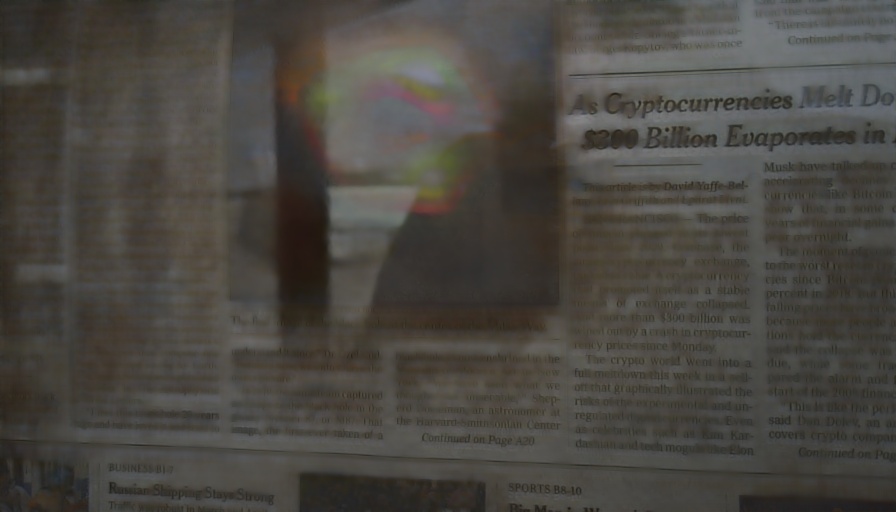} 
                                \\
		(a) Input $I_{a}$                        & (b) Ours  without align.~\cite{Lei_2021_RFC} $I_{fo}$                   & (c) Ours without align.~\cite{Lei_2021_RFC}  $\hat T$     & (d) Ours without align.~\cite{Lei_2021_RFC} $\hat R$   \\
  		\includegraphics[width=0.243\linewidth]{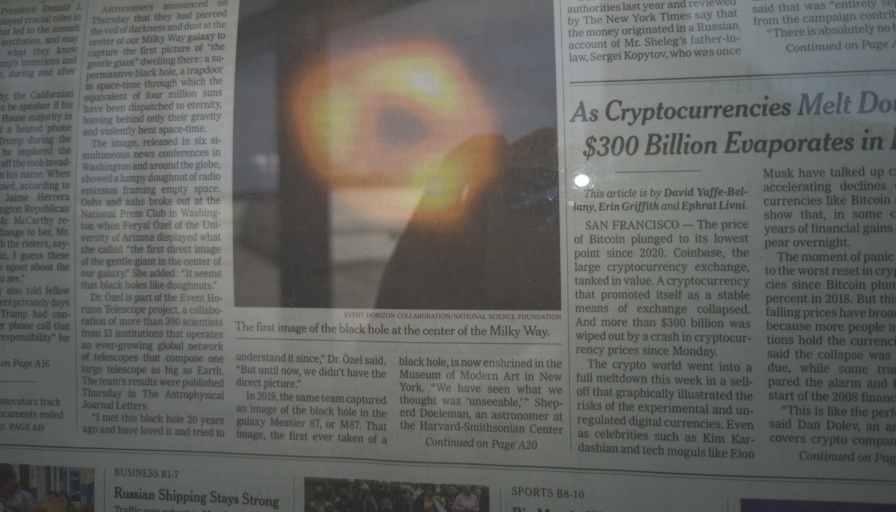}         &
		\includegraphics[width=0.243\linewidth]{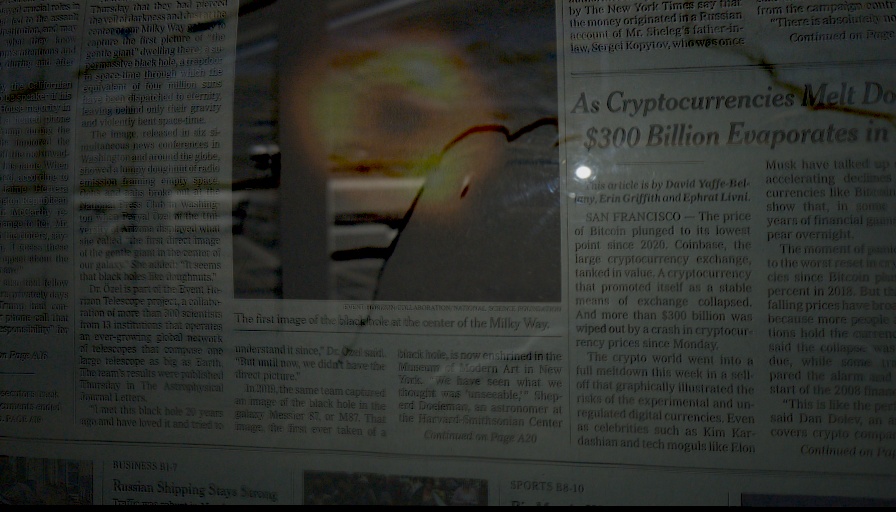}  &                      
		\includegraphics[width=0.243\linewidth]{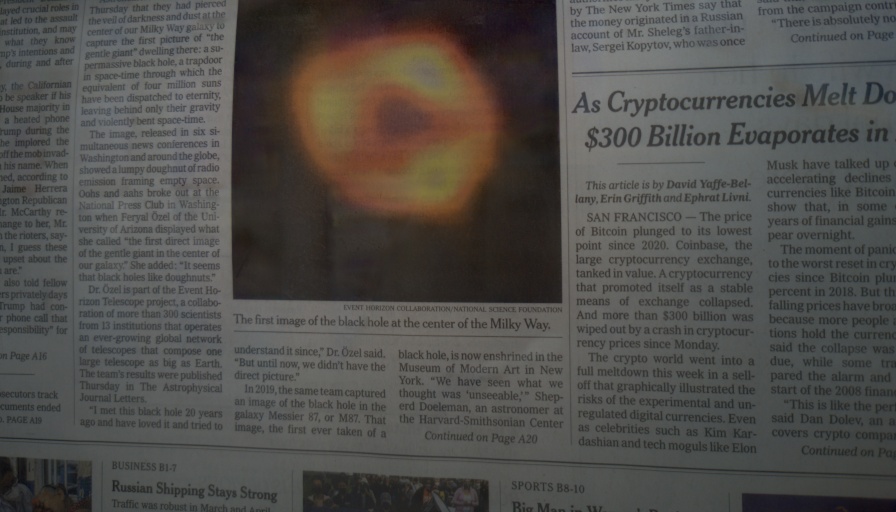} &

  \includegraphics[width=0.243\linewidth]{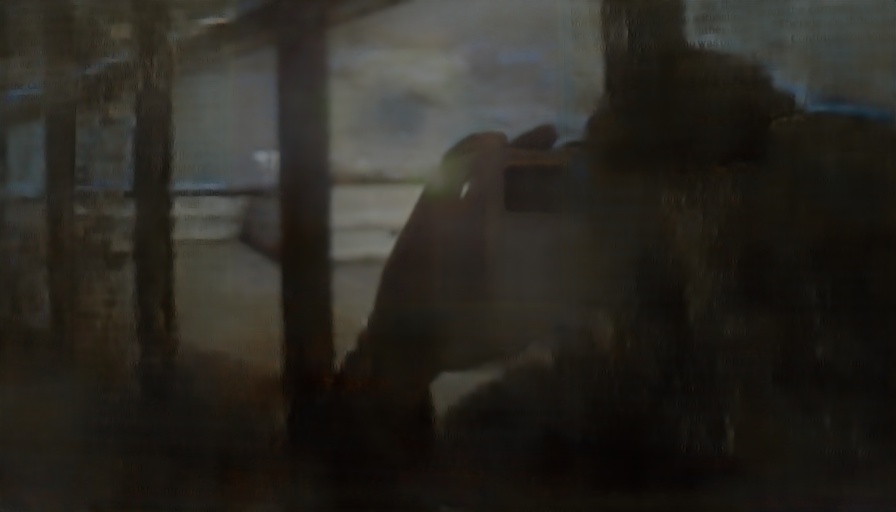} 
		\\
		(e) Input $I_{f}$   & (f)   Ours with align. $I_{fo}$                        
   &(g) Ours  with align. $\hat T$   & (h)     
  Ours  with align. $\hat R$\\
	\end{tabular}

	\vspace{1mm}
	\caption{Qualitative comparison to our previous version without alignment~\cite{Lei_2021_RFC} on real-world handheld images. Without alignment, our previous model suffers from artifacts in both transmission and reflection images. Our current model aligns the transmission layer, so artifacts only appear at the moving edges of the reflection, and the reflection can be removed correctly in the final result. }
	\label{fig:AlignmentPrev}
\end{figure*}

\begin{figure}[]
	\centering
	\begin{tabular}{@{}*{2}{c@{\hspace{1mm}}}c@{}}
		\includegraphics[width=0.45\linewidth]{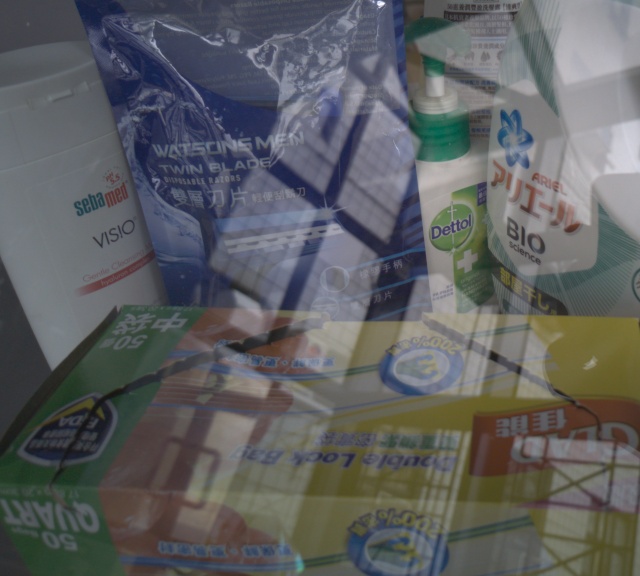} &
		\includegraphics[width=0.45\linewidth]{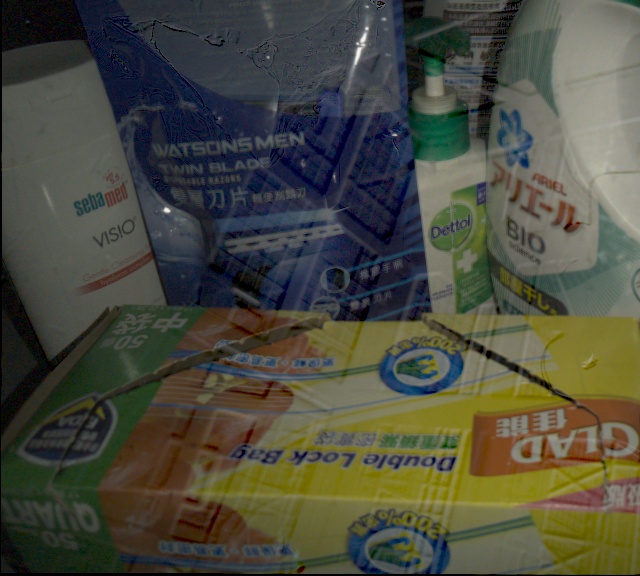}  \\
  Input $I_{a}$                    &  Input aligned $I_{fo}$ \\

		\includegraphics[width=0.45\linewidth]{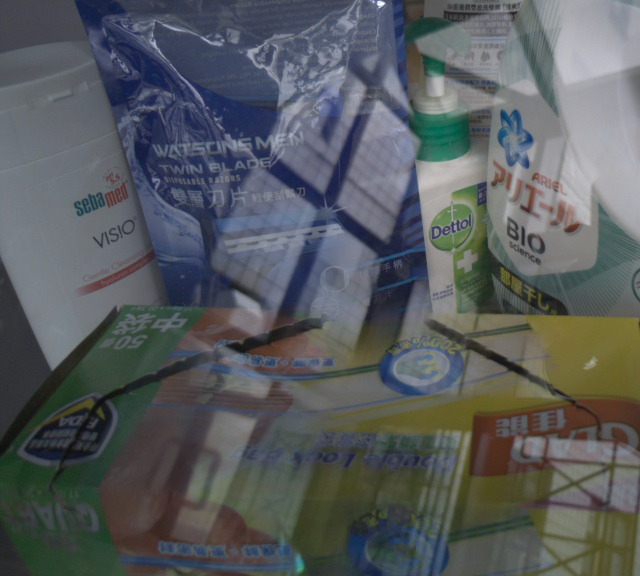} &
  		\includegraphics[width=0.45\linewidth]{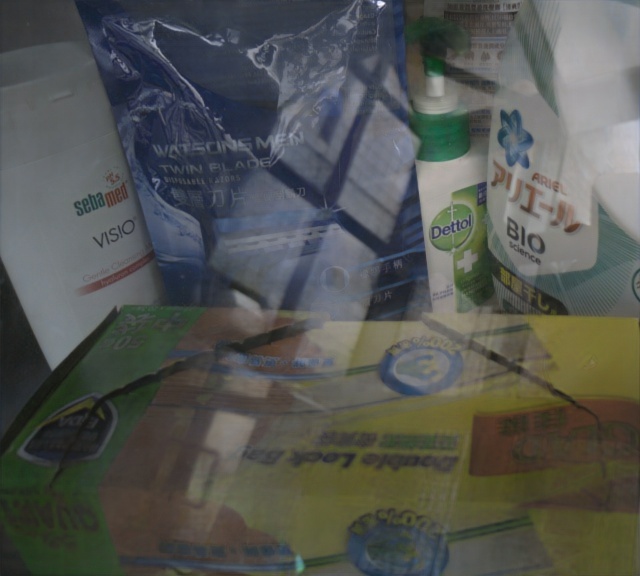} \\
	Zheng et al.~\cite{dong2020location}  & Kim et al.~\cite{Kim_2020_CVPR}    \\
		\includegraphics[width=0.45\linewidth]{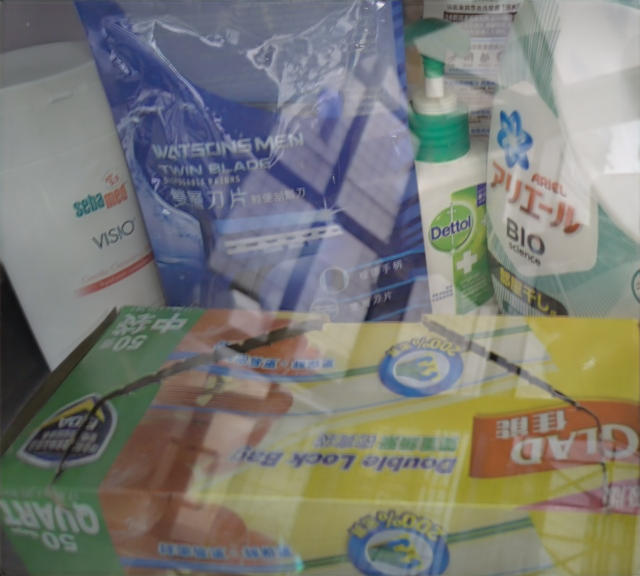}&
  		\includegraphics[width=0.45\linewidth]{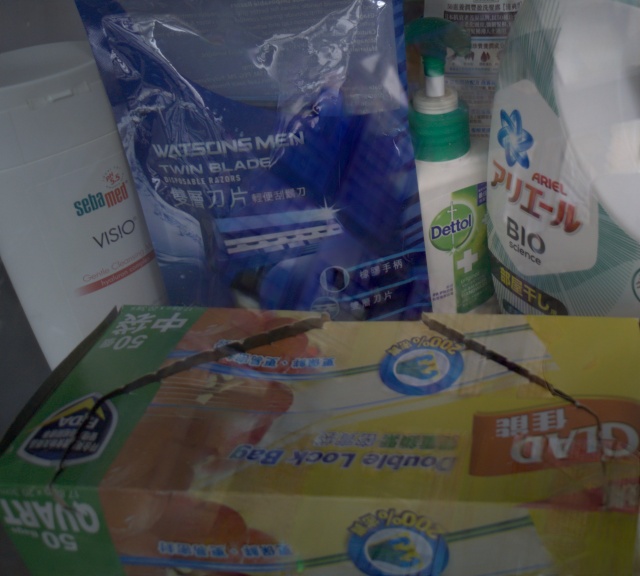}                      \\

		\\
		
	 SDN~\cite{chang2020siamese}                &

  		 Ours $\hat T$   \\

	\end{tabular}
	\vspace{1mm}
	\caption{Qualitative comparison to single-image~\cite{Kim_2020_CVPR,dong2020location} and flash-based~\cite{chang2020siamese} baselines on real-world handheld images. On misaligned handheld images, our model with alignment still can remove the strong reflection that other baselines cannot correctly remove.  }
	\label{fig:AlignBaselines}
\end{figure}

\begin{figure*}[h]
	\centering
	\begin{tabular}{@{}*{3}{c@{\hspace{1mm}}}c@{}}
		\includegraphics[width=0.243\linewidth]{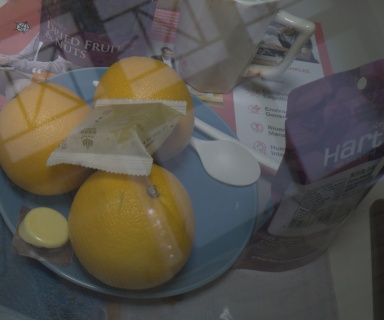}         &
		\includegraphics[width=0.243\linewidth]{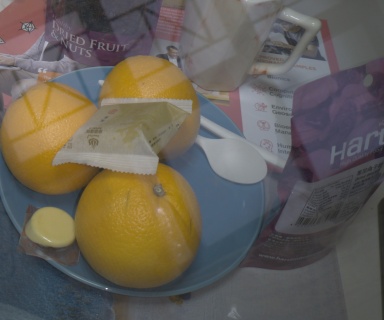}           &
		\includegraphics[width=0.243\linewidth]{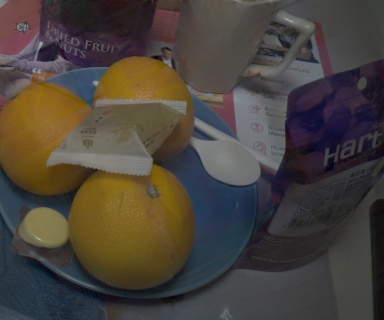} &  
		\includegraphics[width=0.243\linewidth]{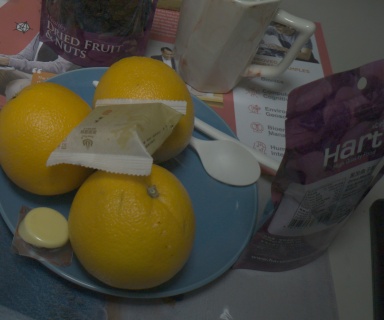}                                                                                                      \\
		(a) Input $I_{a}$                                                                                             & (b) Input $I_{f}$                   & (c) Ours complete  & (d)  Ground truth  \\
		\includegraphics[width=0.243\linewidth]{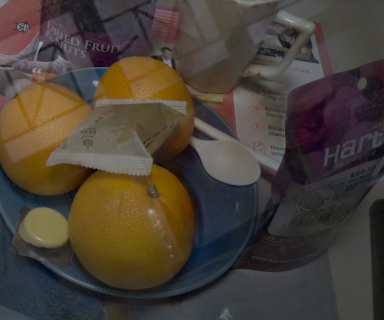} &
		\includegraphics[width=0.243\linewidth]{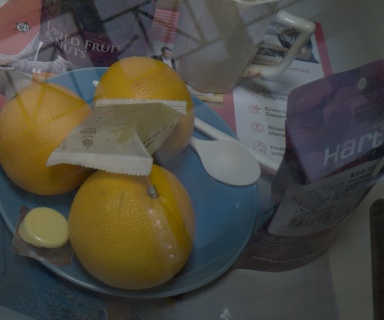} &
		\includegraphics[width=0.243\linewidth]{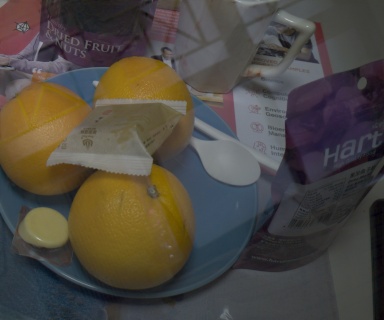} &
		\includegraphics[width=0.243\linewidth]{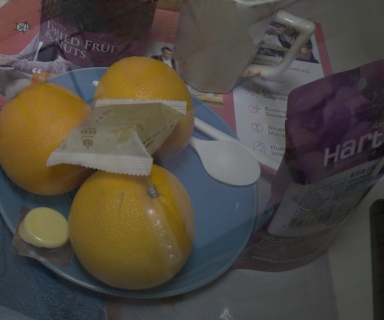}  
		\\
		(e) Without misaligned data                                                                       & (f) Without alignment   &
		(g)  With misaligned data                                                                      & (h) With alignment module\\
			 and alignment module~\cite{Lei_2021_RFC}	&module&(Homography-based) & (Homography-based) 
	\end{tabular}
	\vspace{1mm}
	\caption{Qualitative comparison among our complete model and multiple ablated models on \textcolor{black}{FOR-Misaligned dataset}. When our previous model is trained on misaligned data, there are fewer artifacts in the transmission layer. When random homography data or homography alignment is used, the performance is degraded because homography can only align part of the image if the misalignment is complex. }
	\label{fig:AlignmentAblation}
\end{figure*}

\begin{table}[]
	\small
	\caption{Quantitative comparison to single-image~\cite{Li_2020_CVPR,Kim_2020_CVPR,dong2020location},  flash-based~\cite{chang2020siamese} baselines, and our previous version without alignment~\cite{Lei_2021_RFC} on \textcolor{black}{FOR-Misaligned} dataset. With misaligned training data and the alignment module, we achieved competitive performance on misaligned data. }
	\label{table:CmpToPrev}
	\centering
	\renewcommand{\arraystretch}{1.2}
	\begin{tabular}{l@{\hspace{6mm}}lc@{\hspace{4mm}}c@{\hspace{4mm}}c}
		\toprule[1pt]

		Method      & PSNR$\uparrow$    & SSIM$\uparrow$ &LPIPS$\downarrow$    \\
		\midrule
	 Li et al.~\cite{Li_2020_CVPR} 	&24.76&0.909&\textcolor{black}{0.118}\\
 Kim et al.~\cite{Kim_2020_CVPR}  &24.53&0.911&\textcolor{black}{0.115}\\
 Zheng et al.~\cite{dong2020location}     &22.81&0.913&\textcolor{black}{0.119}\\
 SDN~\cite{chang2020siamese}  &23.44&0.873&\textcolor{black}{0.159}\\
	Ours without align.~\cite{Lei_2021_RFC}        &	25.41&0.917&\textcolor{black}{0.112}\\
Ours with align.       &\textbf{28.60}&\textbf{0.956}&\textbf{\textcolor{black}{0.071}}\\
		\bottomrule[1pt]
	\end{tabular}
	\vspace{1mm}
\end{table}

\begin{table}[t!]
	\centering
		\renewcommand{\arraystretch}{1.2}
 	\small
	\caption{\textcolor{black}{Quantitative comparison about alignment module and misaligned training data on FOR-Misaligned dataset. Both proposed misaligned dataset and designed alignment module can improve the performance on real-world misaligned data. Test-time alignment: align the ambient and flash image at test time.} }

\begin{tabular}{l*{4}{c}}
		\toprule[1pt]

Training data &Alignment & PSNR$\uparrow$ & SSIM$\uparrow$ &LPIPS$\downarrow$  \\
     \hline
     Aligned data  & No  &25.41&0.917&0.112\\
     Aligned data  & Test-time &26.50&0.945&0.087  \\
 Misaligned data   & No  &27.04 &0.940&0.090\\
Misaligned data &   Yes&\textbf{28.60}&\textbf{0.956}&\textbf{0.071}\\
		\bottomrule[1pt]
\end{tabular}
	\label{table:OverallMisalign}
\end{table}

\begin{table*}[t]
	\small
	\caption{Quantitative comparison between different training data synthesis methods on FOR-Misaligned dataset. The use of depth-based misaligned data results in better performance compared to using homography-based data }
	\label{table:TypesOfAligned}
	\centering
		\renewcommand{\arraystretch}{1.2}

\begin{tabular}{p{55mm}*{6}{c}}
		\toprule[1pt]

Misaligned training data&\multicolumn{6}{c}{Alignment module}  \\

synthesis method&\multicolumn{3}{c}{Without alignment}&  \multicolumn{3}{c}{With alignment}  \\

 & PSNR$\uparrow$ & SSIM$\uparrow$&LPIPS$\downarrow$  &  PSNR$\uparrow$ & SSIM$\uparrow$&LPIPS$\downarrow$   \\
     \hline
   None (aligned data) & 25.41& 0.917& 0.112 &  26.49&	0.942	&0.088  \\
   Homography-based misaligned data &  26.46&0.938&0.099&  27.58&0.950&\textcolor{black}{0.081}\\
Depth-based misaligned data &\textbf{27.04}&\textbf{0.940}&\textbf{0.090}&\textbf{28.60}&\textbf{0.956}&\textbf{\textcolor{black}{0.071}}\\
		\bottomrule[1pt]

\end{tabular}

	\vspace{1mm}
\end{table*}

\begin{table}[]
  \small
  \caption{Quantitative comparison between alignment methods on FOR-Misaligned dataset. All models are trained on the misaligned data for fair comparison. The performance benefits from the expressive capability of optical flow estimation network better when training data with more complex misalignment is applied. }
  \label{table:Alignment}
  \centering
  \renewcommand{\arraystretch}{1.2}

\begin{tabular}{l*{3}{c}}
    \toprule[1pt]
Alignment module& PSNR$\uparrow$ & SSIM$\uparrow$ & LPIPS$\downarrow$  \\
     \hline
\textcolor{black}{No alignment}  & 27.04&0.940&0.090 \\
Homography  & 27.97&0.952& 0.072\\
Optical Flow (pretrained)  &  28.19&0.954&0.073\\
Optical Flow (finetuned)& \textbf{28.60}&\textbf{0.956} & \textbf{0.071}\\
    \bottomrule[1pt]
  
\end{tabular}
  \vspace{1mm}
\end{table}

\begin{figure}[h]
	\centering
	\begin{tabular}{@{}*{2}{c@{\hspace{1mm}}}c@{}}
		\includegraphics[width=0.321\linewidth]{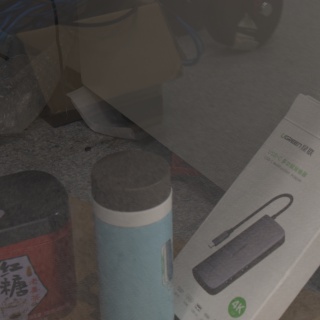}         &
		\includegraphics[width=0.321\linewidth]{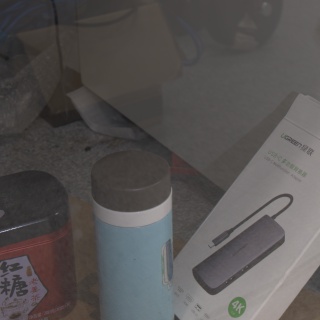}           &
		\includegraphics[width=0.321\linewidth]{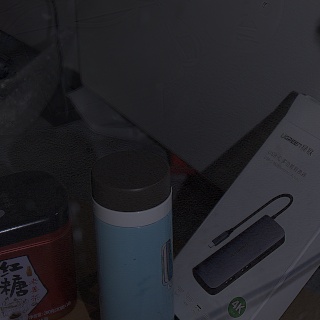}        \\
		 Input $I_{a}$                                                                                             &  Input $I_{f}$                   &  Processed $I_{fo}$  \\
\includegraphics[width=0.321\linewidth]{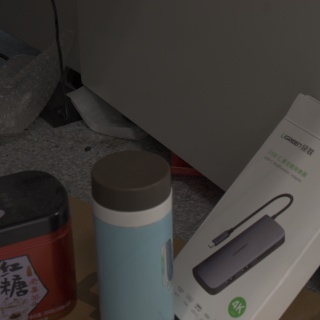} &
		\includegraphics[width=0.321\linewidth]{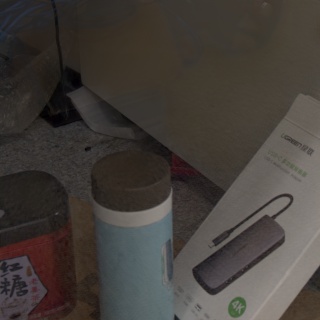} &
		\includegraphics[width=0.321\linewidth]{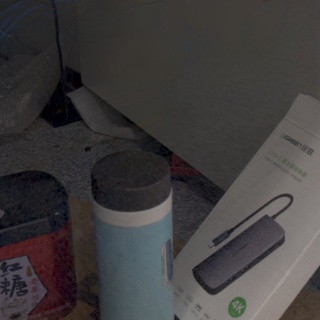}
		\\ Ground truth &
		 $\hat T$ with sRGB                                                                      &  $\hat T$ with linear RGB     \\
	\end{tabular}
	\vspace{1mm}
	\caption{Qualitative comparison between using sRGB images and linear RGB images on aligned data. The difference between using linear RGB images and sRGB images is minor. }
	\label{fig:PreImages}
\end{figure}
\subsection{Handling Misalignment in Handheld Photography}
In this section, we analyze the effectiveness of our proposed pipeline in Section~\ref{subsec:misalign} for the misalignment between the flash and no-flash images. We conduct the experiments on our collected FOR-Misaligned dataset as introduced in Section~\ref{subsec:misalign_data}. 

\subsubsection{Comparison to Baselines}
We first compare our approach with the previous version on misaligned data. As shown in Table~\ref{table:CmpToPrev}, our current model outperforms the previous method~\cite{Lei_2021_RFC} significantly on \textcolor{black}{quantitative test data in FOR-Misaligned dataset}. Figure~\ref{fig:AlignmentPrev} shows an example of a real-world handheld flash and no-flash image pair \textcolor{black}{from FOR-Misaligned qualitative evaluation data}. If we directly compute the flash-only image without alignment, there exist artifacts in both the transmission and reflection part of the flash-only image. The intermediate reflection prediction is corrupted by ghosting transmission resulting from the imperfect flash-only transmission, which lets transmissions be removed instead of the reflection in the final transmission prediction. By aligning the pair before computing the flash-only image, we can get a clear static flash-only transmission with some artifacts only on the moving edges of reflection. The resulting reflection prediction is cleaner, and the final transmission prediction can remove reflection correctly. Table~\ref{table:CmpToPrev} also shows that our aligned model outperforms other single-image~\cite{Li_2020_CVPR,Kim_2020_CVPR,dong2020location} and flash-based~\cite{chang2020siamese} baselines. Figure~\ref{fig:AlignBaselines} shows \textcolor{black}{another qualitative result on FOR-Misaligned dataset}. With misaligned training data and the alignment module, our approach can still remove the strong reflection that other baselines fail to remove.

\subsubsection{Controlled experiments} 
We provide controlled experiments to analyze the importance of each designed module. \textcolor{black}{Table~\ref{table:OverallMisalign} shows that both the simulated misaligned data and the alignment module can improve the performance on real-world misaligned data.}

\textbf{Misaligned training data.} In Table~\ref{table:TypesOfAligned}, we present the results on different types of training data. In addition to the misaligned data synthesized by depth, we also implement a baseline using randomly generated homography. Firstly, we show that better misaligned training data can directly improve the performance of the previous framework (i.e., without alignment module~\cite{Lei_2021_RFC}). Besides, for our current version (i.e., with the alignment module), we also observe consistent improvement with better training data. Figure~\ref{fig:AlignmentAblation} presents the perceptual results for different models. The result obtained by our depth-based synthesis pipeline achieves much better performance than the other ablated models.

\textbf{Alignment.} Aligning the flash and ambient images can improve the performance further. As shown in Table~\ref{table:Alignment}, we implement different ablated models to analyze the effectiveness of our approach. 
Specifically, we tried using SIFT to estimate the homography and using optical flow to obtain dense correspondences. All methods that use the alignment module can improve performance. In addition, using optical flow can get better performance than using homography as homography cannot model complex misalignment. Finetuning the PWC-Net on the reflection dataset can improve the performance further, and we believe it is because the pretrained network is not capable of handling images with strong reflection.

Figure~\ref{fig:AlignmentAblation} shows the qualitative results. Due to the expressiveness limitation of homography, for complex misalignment, only part of the image can be aligned. Strong reflection still cannot be fully removed on other misaligned parts, as shown in Figure~\ref{fig:AlignmentAblation}(h).

\begin{table}[]
	\small
	\caption{Quantitative comparison the relative performance of the two models in different image space. The computation of flash-only images is critical for sRGB images, but it is not required for linear images. Note that the training data uses a simple ISP compared with previous version. }
	\label{table:PreImages}
	\centering
	\renewcommand{\arraystretch}{1.2}
	\begin{tabular}{l@{\hspace{6mm}}c@{\hspace{4mm}}c@{\hspace{4mm}}c}
		\toprule[1pt]

		Input           & PSNR$\uparrow$    & SSIM$\uparrow$& LPIPS$\downarrow$     \\
		\midrule
		Linear Flash &30.88 &0.938&\textcolor{black}{0.090}\\
Linear Flash-only&30.86 &0.937&\textcolor{black}{0.089}\\
\hline
sRGB Flash  &30.54 &0.923&\textcolor{black}{0.091}\\
sRGB Flash-only  &  {31.30} & {0.944}&{\textcolor{black}{0.084}}\\
		\bottomrule[1pt]
	\end{tabular}
	\vspace{1mm}
\end{table}
\subsection{Analysis for Linear RGB Images}

\textcolor{black}{The motivation of this experiment is detailed in Section~\ref{subsec:linear}. Briefly speaking, the use of human-designed reflection-free flash-only images is key to the success of the method. It is reasonable since we cannot compute a flash-only image \textcolor{black}{easily} using a pair of sRGB images since the linearity does not hold on sRGB space and it is usually hard to convert the sRGB images to raw images. In this section, we further analyze the performance of using linear RGB images, which have the linearity property. 
}

\textcolor{black}{Table~\ref{table:PreImages} shows the quantitative results. We implement two controlled models with linear RGB images and sRGB images, respectively. For each type of image, we directly input the flash image or compute the flash-only image. For linear RGB images, we notice that there is almost no improvement by computing the flash-only image. We show that by using raw images where linearity holds, a flash-only image does not need to be manually computed for good performance. However, for the sRGB images, computing the flash-only image as the input can effectively improve the performance. This fact reveals that the linearity property is a key property to success. In addition, we notice that using linear images cannot improve the quantitative results on sRGB space. It might be caused by the domain gap between sRGB space and linear RGB space. Figure~\ref{fig:PreImages} shows the qualitative results. The difference between linear RGB images and sRGB images is minor. Note that since the experimental setup is different from Table~\ref{table:Ablation study}, the results are different: the relative improvement of using the RGB flash-only image is fewer since the ISP is simpler in this case. 
}

\section{Conclusion}
We propose a very simple yet effective cue called \emph{reflection-free cue} for reflection removal, which is independent of the appearance and strength of reflection. The reflection-free cue is based on the fact that objects in reflection do not directly receive light from the flash and the reflected flash is weak. With a reflection-free flash-only image as guidance, estimating the reflection becomes much easier. Since the flash-only image has obvious artifacts, we propose a dedicated architecture to avoid absorbing artifacts of flash-only images and utilize the cue better. As a result, our model outperforms state-of-the-art methods significantly on a real-world dataset. Also, the qualitative results show that our method can robustly remove various kinds of reflections. We also analyze the flash-based method's feasibility and find it simple to continuously take two images, making it practical in real-world applications. At last, we extend our approach to handheld photography with our dedicated misaligned training data and pre-alignment strategy so that our model can be robust to the misalignment between the flash image and ambient image. We study the performance of using linear images as input and observe that flash-only cues can be learned in the linear space.

\section*{Acknowledgements}
\addcontentsline{toc}{section}{Acknowledgements}This work was supported by the InnoHK program. We thank Xuaner Zhang, Changlin Li, Yazhou Xing, and anonymous reviewers for helpful discussions on the paper.


\ifCLASSOPTIONcaptionsoff
	\newpage
\fi



%\bibliographystyle{IEEEtran}

\bibliographystyle{IEEEtran}
\bibliography{reference}



% \end{thebibliography}

\vspace{-6mm}

\begin{IEEEbiography}[{\includegraphics[width=1in,height=1.25in,clip,keepaspectratio]{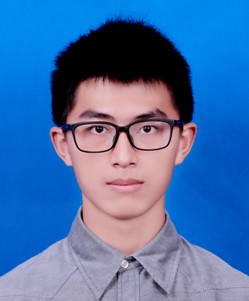}}]{Chenyang Lei} is an assistant professor at the Centre for Artificial Intelligence and Robotics (CAIR), Chinese Academy of Sciences. He is also a research scholar at Princeton University. He received his Ph.D. in computer science from the HKUST in 2022, and a B.E. degree from Zhejiang University in 2018. He feels interested in computation
	photography, low-level computer vision, and video processing.
\end{IEEEbiography}

\vspace{-6mm}

\begin{IEEEbiography}[{\includegraphics[width=1in,height=1.25in,clip,keepaspectratio]{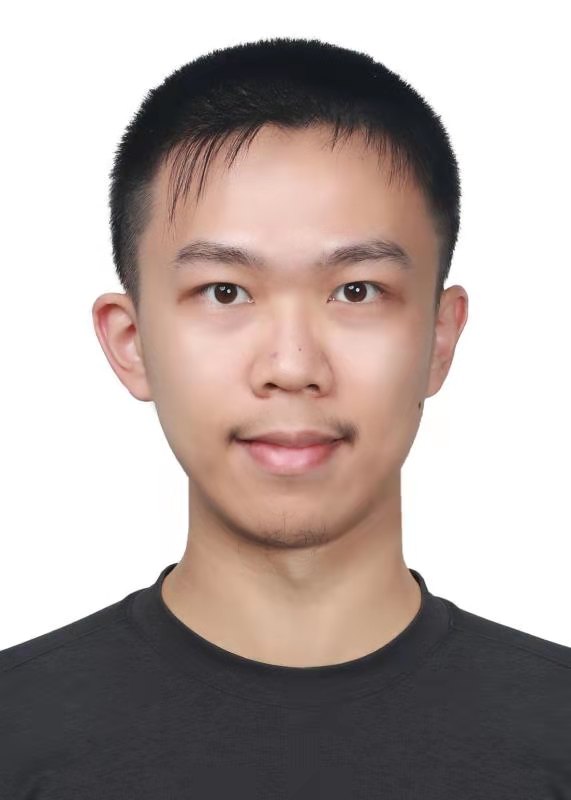}}]{Xudong Jiang}
	is currently pursuing an MSc degree at ETH Zurich. He feels interested in computation
	photography, low-level computer vision, and video processing.
\end{IEEEbiography}

\vspace{-6mm}

\begin{IEEEbiography}
	[{\includegraphics[width=1in,height=1.25in,clip,keepaspectratio]{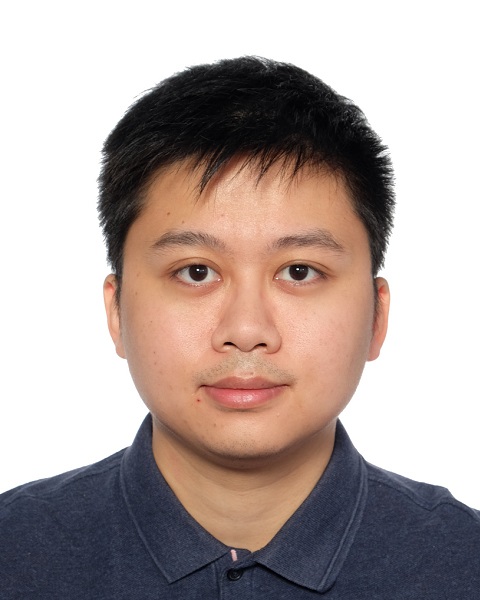}}]
	{Qifeng Chen} is an assistant professor of the Department of Computer Science and Engineering and the Department of Electronic and Computer Engineering at HKUST. He received his Ph.D. in computer science from Stanford University in 2017, and a bachelor's degree in computer science and mathematics from HKUST in 2012. He is named one of 35 Innovators under 35 in China in 2018 by MIT Technology Review. He won the Google Faculty Research Award 2018. He is the HKUST ACM programming faculty coach and won the 2nd place worldwide at the ACM-ICPC World Finals in 2011. He is a member of IEEE.
\end{IEEEbiography}

\end{document}